\documentclass[acmtog,nonacm]{acmart}
\newcommand{\HL}[1]{\iftrue {\color{red}[HL: #1]}\else {}\fi} 
\newcommand{\EZ}[1]{\iftrue {\color{blue}[EZ: #1]}\else {}\fi}
\newcommand{\fong}[1]{\iftrue {\color{yellow}[Fong: #1]}\else {}\fi}
\newcommand{\anh}[1]{\iftrue {\color{cyan}[Anh: #1]}\else {}\fi}

\AtBeginDocument{%
  }

\usepackage[capitalize]{cleveref}
\usepackage{multirow}
\usepackage{tabularx}

\DeclareMathOperator*{\argmin}{arg\,min}

\begin{document}

\title{VOODOO XP: Expressive One-Shot Head Reenactment for VR Telepresence}

\author{Phong Tran}
\email{the.tran@mbzuai.ac.ae}
\affiliation{%
  \institution{Mohamed Bin Zayed University of Artificial Intelligence}
  \country{UAE}
}

\author{Egor Zakharov}
\affiliation{%
  \institution{ETH Zurich}
  \country{Switzerland}}
\email{ezakharov@ethz.ch}

\author{Long-Nhat Ho}
\affiliation{%
  \institution{Mohamed Bin Zayed University of Artificial Intelligence}
  \country{UAE}
}
\email{long.ho@mbzuai.ac.ae}

\author{Liwen Hu}
\affiliation{%
 \institution{Pinscreen}
 \country{USA}}
\email{liwen@pinscreen.com}

\author{Adilbek Karmanov}
\affiliation{%
  \institution{Mohamed Bin Zayed University of Artificial Intelligence}
  \country{UAE}
}
\email{adilbek.karmanov@mbzuai.ac.ae}

\author{Aviral Agarwal}
\affiliation{%
 \institution{Pinscreen}
 \country{USA}}
\email{aviral@pinscreen.com}

\author{McLean Goldwhite}
\affiliation{%
 \institution{Pinscreen}
 \country{USA}}
\email{lain@pinscreen.com}

\author{Ariana Bermudez Venegas}
\affiliation{%
  \institution{Mohamed Bin Zayed University of Artificial Intelligence}
  \country{UAE}
}
\email{ariana.venegas@mbzuai.ac.ae}

\author{Anh Tuan Tran}
\affiliation{%
 \institution{VinAI Research}
 \country{Vietnam}}
\email{anhtt152@vinai.io}

\author{Hao Li}
\affiliation{%
 \institution{Pinscreen, MBZUAI}
 \country{USA}}
\email{hao@hao-li.com}

\renewcommand{\shortauthors}{Tran et al.}

\begin{abstract}
We introduce VOODOO XP: a 3D-aware one-shot head reenactment method that can generate highly expressive facial expressions from any input driver video and a single 2D portrait. 
Our solution is real-time, view-consistent, and can be instantly used without calibration or fine-tuning. We demonstrate our solution on a monocular video setting and an end-to-end VR telepresence system for two-way communication. Compared to 2D head reenactment methods, 3D-aware approaches aim to preserve the identity of the subject and ensure view-consistent facial geometry for novel camera poses, which makes them suitable for immersive applications. While various facial disentanglement techniques have been introduced, cutting-edge 3D-aware neural reenactment techniques still lack expressiveness and fail to reproduce complex and fine-scale facial expressions.
We present a novel cross-reenactment architecture that directly transfers the driver's facial expressions to transformer blocks of the input source's 3D lifting module. We show that highly effective disentanglement is possible using an innovative multi-stage self-supervision approach, which is based on a coarse-to-fine strategy, combined with an explicit face neutralization and 3D lifted frontalization during its initial training stage. We further integrate our novel head reenactment solution into an accessible high-fidelity VR telepresence system, where any person can instantly build a personalized neural head avatar from any photo and bring it to life using the headset. We demonstrate state-of-the-art performance in terms of expressiveness and likeness preservation on a large set of diverse subjects and capture conditions.
\end{abstract}




\begin{teaserfigure}
\includegraphics[width=1.0\textwidth]{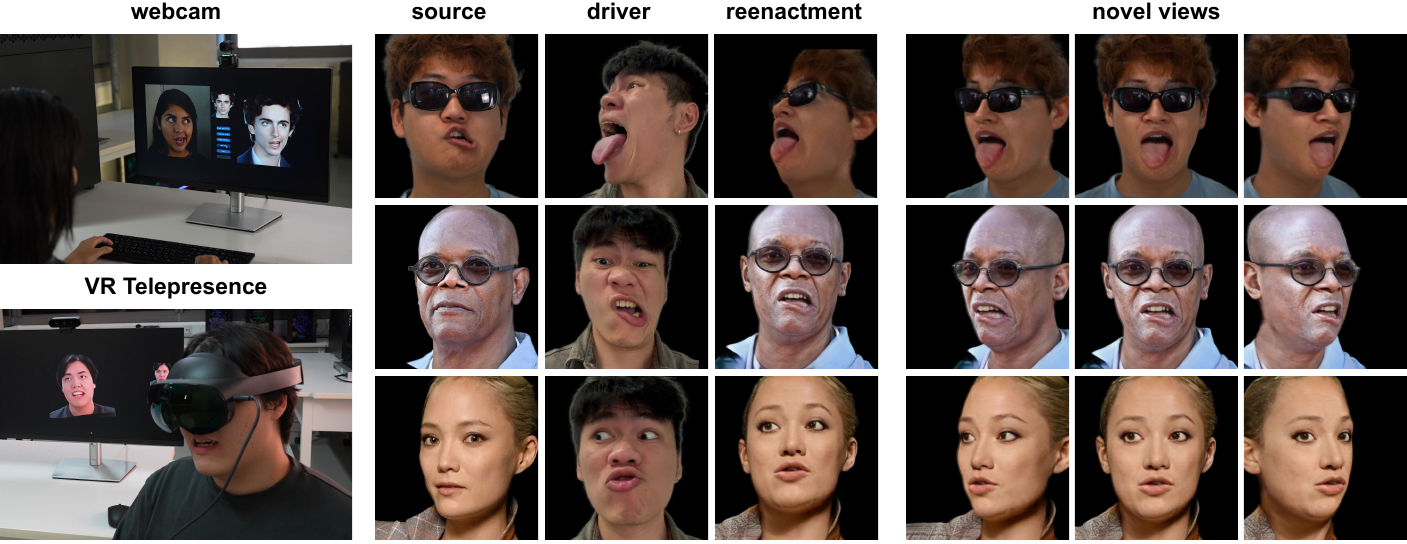}
\caption{We introduce \textbf{VOODOO XP}, an expressive 3D-aware head reenactment system. Our method successfully transfers the driver's expressions to the source and produces view-consistent renderings, covering a wide range of head poses and highly dynamic and extreme expressions from the input driver. Thanks to the real-time capability of our method, we also demonstrate our solution in a live monocular webcam setting and an end-to-end VR telepresence system for two-way communication. Please see the supplement video for the live demonstration.}
\label{fig:teaser}
\end{teaserfigure}

\maketitle

\section{Introduction}
Virtual and augmented reality has the potential to displace traditional 2D video conferencing and facilitate fully immersive 3D communication in virtual worlds and remote spatial collaboration. The latest consumer AR/VR headsets (e.g., Apple’s Vision Pro~\cite{applevisionpro}, Meta Quest Pro~\cite{metaquestpro}), all support built-in real-time face tracking capabilities~\cite{li2015facial, olszewski2016high, DBLP:journals/corr/abs-1808-00362, Ma_2021_CVPR, 10.1145/3306346.3323030}, which can enable engaging social face-to-face interactions using 3D avatars. Those avatars are typically built as a separate process without the user wearing the HMD using an authoring tool, from scans or videos swiped around a face, or by uploading a photo. While most VR applications (e.g., Meta Horizon) are based on conventional graphics rendering of 3D avatars, Apple’s Persona uses a 2D neural rendering approach and a face scanning procedure to make the creation of realistic avatars easier. 

Despite the steady progress of real-time game engines, such as Unreal~\cite{unrealengine} and Unity~\cite{unityengine}, and their capabilities of rendering highly quality digital humans (e.g., MetaHuman Creator~\cite{metahuman}), achieving realism for personalized avatars, using such graphics pipeline, typically relies on professional scans, complex facial rig adjustment, and careful shader/lighting tweaks for individual scenes. On the other hand, neural rendering approaches are gaining popularity as they ensure the generation of photorealistic faces and expressions using models trained directly from real world data. High-end neural rendering systems such as Meta’s avatar codec model~\cite{DBLP:journals/corr/abs-1808-00362, Ma_2021_CVPR, 10.1145/3306346.3323030} a personalized head avatar using a complex multi-view video capture system or a number of 3D facial scans obtained from a handheld device~\cite{CaoChen2022authentic}. The process, however, generally involves tedious recording sessions, as well as a lengthy optimization process (hours to days), which makes it not yet suitable for deployment in everyday environments.

A number of one-shot neural head reenactment methods have shown that high-fidelity facial expressions and poses can be generated instantly from a single portrait using a driver video by training a generative model using large image/video datasets~\cite{Wiles18,Siarohin_2019_CVPR,siarohin2019first,zakharov2019few,burkov2020neural,zakharov2020fast,doukas2020headgan,song2021pareidolia,wang2021latent,wang2021one,Ren_2021_ICCV,siarohin2021motion,EAMM_2022,hong2022depth,Tao_2022_CVPR,StyleHEAT_2022,drobyshev2022megaportraits,Zhao_2022_CVPR,Zhang_2023_CVPR,Gao_2023_CVPR,xiang2020one}. However for real-time 2D methods, the identity and expressions can deviate drastically from the input, especially when a novel view is rendered that is different from the driver video, which is the case for VR telepresence. Lately, several 3D-aware head reenactment methods have been introduced and use neural radiance fields combined with a disentanglement approach for identity, expressions, and pose to achieve consistent renderings from novel views~\cite{Schwarz2020graf,Chan_2021_CVPR,Niemeyer_2021_CVPR,chan2022efficient,Or-El_2022_CVPR,epigraf2022,Xue_2022_CVPR,Deng_2022_CVPR,Xiang_2023_ICCV,An_2023_CVPR,xu2022pv3d,Xu_2023_CVPR,li2023generalizable,tran2023voodoo, ki2024learning, ye2024real3dportrait}. Due to the volumetric nature of neural fields, such disentanglement is particularly challenging, and despite the most recent efforts based on either 3D morphable models~\cite{li2023generalizable}, 3D lifted frontalization techniques~\cite{tran2023voodoo}, or multi-modal training-based expression encoder~\cite{deng2024portrait4d,deng2024portrait4dv2}, the reenacted faces struggle with complex face/lip movement, as well as asymmetric expressions.

Expressive faces are important for conveying emotions and producing natural lip movements during speech in social interactions. To this end, we propose a real-time 3D head reenactment approach called VOODOO XP, which can generate highly expressive faces and view-consistent renderings given a single unconstrained input portrait (source), a monocular video performance of the same or different person (driver), as well as an arbitrary camera pose. Additionally, we also introduce the first end-to-end VR telepresence system for two-way face-to-face communication based on a one-shot head reenactment technology and the Meta Quest Pro HMD. The user first takes a picture with a webcam or picks an arbitrary 2D portrait as the source, and after wearing the VR headset, the user can instantly drive the face of its neural head avatar and communicate immersively with a second person. 

Similar to the state-of-the-art neural head reenactment approach, VOODOO 3D~\cite{tran2023voodoo}, our method performs a fully volumetric disentanglement for identity, expressions, and pose using a 3D lifted frontalization network. However, instead of 3D lifting the driver's face and extracting its expressions, we directly transfer the expression signal to transformer blocks of the 3D lifting module used for the source portrait. For an effective disentanglement and extraction of the driver's expression, we present a multi-stage self-supervised learning approach, in which the first stage uses an explicit neutralization and 3D lifted frontalization for training, followed by less restrictive but higher resolution fine-scale training, and global fine-tunine step. We demonstrate extreme facial expressions such as face squeezing, complex lip movements, asymmetric facial poses, and fine-scale wrinkles that are particularly challenging for other methods. It is worth noting that our method can synthesize tongues as well as correct eyebrow movements behind glass frames. 

While we proposed a one-shot technique, our system naturally supports few-shot fine-tuning as an optional step. In case additional sources images are available, more faithful avatars can be obtained in tens of seconds. We show how our system achieves superior reenactment in terms of expressiveness, fidelity, and identity preservation compared to the current state-of-the-art on various datasets and a wide range of examples. 
We also demonstrate the performance and robustness of our one-shot reenactment technique on an integrated, immersive telepresence system for two-way communication using VR headsets. We make the following contributions:

\begin{itemize}
    \item We introduce a novel 3D-aware neural head reenactment architecture using a transformer-based expression transfer approach for generating highly expressive facial expressions in real-time.
    \item For effective volumetric disentanglement of complex expressions, we propose a multi-stage training approach based on face neutralization and 3D lifted frontalization, coarse-to-fine training, and global fine-tuning.
    \item We present the first end-to-end VR telepresence solution based on a one-shot 3D head reenactment algorithm.
\end{itemize}

\section{Related work}

For today's VFX, gaming, and VR/AR applications, the standard for rendering 3D avatars consists of using a conventional graphics pipeline based on textured mesh representations, followed by shader and lighting processing. Personalized avatar heads are either created from scratch by skilled artists and/or assisted using sophisticated authoring tools for digital humans (e.g., Metahuman Creator~\cite{metahuman}). 3D scanning techniques based on multi-view or photometric stereo are often used to capture accurate geometry and skin appearance properties and facilitate the creation of high-fidelity digital humans, but they require a complex capture setup. More deployable solutions such as using handheld monocular cameras or depth sensors exist, but achieving photorealism is often associated with a tedious shader/lighting fine-tuning process, and custom modeling approaches for face regions, hair, and eyes are typically required and remain challenging for real-time performance. 

\paragraph{3D Neural Head Modeling from Multi-View Stereo.}

Deep learning-based (neural) rendering techniques are becoming increasingly popular as they facilitate the generation of photorealistic avatars using implicit representations learned directly from captured data (images, videos, or 3D scans). A comprehensive survey can be found in~\cite{DBLP:journals/corr/abs-2004-03805}. Such a holistic approach enables a unified way of modeling and rendering every head component, including hair, eyes, and glasses, including view-dependent appearances and complex material properties. In particular, there is no need for developing specialized models or shaders, and photorealistic 3D avatars can be achieved more easily and without the need for human labor, which is appealing for lifelike VR telepresence experiences. For high-end use cases and research settings, complex multi-view video camera systems are used to build a comprehensive dynamic neural head model of a person with maximum face coverage.

Meta's Deep Appearance Models, or Codec Avatars~\cite{DBLP:journals/corr/abs-1808-00362}, use a 3D mesh representation combined with view-dependent dynamic textures, represented by a variational autoencoder (VAE) network, for realistic rendering of a head avatar. Their solution also integrates inward-facing HMD cameras for real-time avatar animation in a VR setting. 
%
Later works enhance the facial expressiveness of the avatars \cite{10.1145/3306346.3323030} or develop a lightweight model for running on a mobile VR headset ~\cite{Ma_2021_CVPR}. However, for all Codec Avatar methods, a dense multi-view stereo system (40 cameras) is used for learning dynamic facial textures per subject, and a coarse triangle mesh is needed for real-time speed.

Lombardi and colleagues~\cite{Lombardi:2019} introduced a fully volumetric approach called Neural Volumes, which uses multiple views of a dynamic object as input to an encoder-decoder architecture. The latent code generates dense 3D voxels that represent appearance and opacity, which are sampled using a differentiable ray marching algorithm and used for end-to-end training. This approach is suitable for rendering complex volumetric structures like hair and transparent objects but suffers from an extensive memory footprint. 
Mixture of Volumetric Primitives ~\cite{DBLP:journals/corr/abs-2103-01954} addressed the cubic memory requirement problem using a sparse 3D data structure based on rigidly-moving volumetric primitives, initialized with a coarse tracked mesh, to model deformable objects and produced higher quality renderings. Garbin et al.~\cite{garbin2022voltemorph} instead used an efficient deformable tetrahedral graph for real-time rendering of generic dynamic objects in a multi-view setting. The authors presented drivable neural head avatars for VR telepresence applications using the volumetric tetrahedral structure as an extension to a tracked facial blendshape model~\cite{li2017learning}. Similar to the mentioned mesh-driven neural rendering methods, a complex multi-view stereo scanner is needed, followed by hours of data processing to build an avatar head, limiting its deployability.

Neural Radiance Fields (NeRFs)~\cite{mildenhall2021nerf} enable highly realistic renderings as they learn view-dependent appearances. The original work was designed for static scenes captured from large numbers of calibrated views or a monocular moving video sequence with camera poses obtained from structure-from-motion. NeRF techniques that can handle deformable scenes~\cite{du2021neural,li2021neural,park2021nerfies,park2021hypernerf,pumarola2020d,tretschk2021nonrigid} have been introduced for monocular video input, and are generally based on computing a mapping between a canonical and deformed state using deep neural network. More recently, Kirschstein and coworkers have presented a multi-view NeRF reconstruction and rendering method for human heads called Nersemble~\cite{Kirschstein_2023}. Their system combines a deformation field and multi-resolution hash encoding approach to facilitate sparse (16-view) multi-view video input and efficient high-fidelity rendering with blendshape type controls.

3D Gaussian Splatting~\cite{kerbl20233d} has been recently introduced as an alternative to NeRFs for high-quality radiance field rendering from multiple calibrated views without a deep neural network. The method achieves superior efficiency and fidelity using anisotropic 3D Gaussians as an unstructured representation of radiance fields combined with fast GPU-based differentiable rendering. Qian et al. have recently developed a 3D Gaussian-based Head Avatar system~\cite{qian2023gaussianavatars} using video sequences recorded from 16 views and shown that these 3D Gaussians can be initialized and controlled using a FLAME model \cite{li2017learning} and produce superior facial expression quality and accuracy than NeRF-based solutions. A similar approach was proposed by Xu et al.~\cite{xu2023gaussianheadavatar} but uses two separate 3D Gaussian representations, one for neutral face and one for facial expressions, followed by a super-resolution network to produce 2K image output. Saito et al.~\cite{saito2023relightable} recently demonstrated that relightable Gaussian avatars are possible using learnable radiance transfer functions using the UV space of a parametric mesh model. While multi-view capture approaches for neural head avatar modeling produce the highest fidelity and most accurate results, the complex hardware requirements make them unsuitable for consumer avatar creation and everyday deployment.



    


\paragraph{3D Neural Head Modeling from Monocular Video.}


A more accessible approach for building a neural head avatar consists of recording various facial expressions and poses from a video sequence obtained by a single monocular camera. Deep Video Portraits~\cite{kim2018deepvideo} uses a source video to control the facial performance and head pose of a person in a target video. The method consists of extracting dynamic 3DMM models for both source and target subject and translating the retargeted performance using a rendering-to-video translation network~\cite{DBLP:conf/miccai/RonnebergerFB15}, which is trained on all the frames of the target video. Thies and colleagues~\cite{DBLP:journals/corr/abs-1904-12356} introduced a deferred neural rendering framework where neural texture maps, which are attached to a face rig, are learned jointly, in order to produce more detailed facial expressions.

Gafni and coworkers developed a NeRF-based face reenactment framework~\cite{gafni2021dynamic}, called NerFACE, that can be trained from 2 minutes of video recordings and reenacted by a driver video in real-time. They use a 3DMM and camera parameters as input to a dynamic radiance field network, which is optimized using the monocular video sequence, to synthesize a novel expression and view of a target person. Similarly, Guo et al.~\cite{guo2021adnerf} have shown that audio-driven NeRF-based facial reenactment is possible. Grassal et al.~\cite{Grassal_2022_CVPR} model the head avatar using an explicit mesh representation and neural textures. Similar to NerFACE~\cite{gafni2021dynamic}, training time is very long, with 7 hours on two NVIDIA A100 GPUs. Zheng et al.~\cite{Zheng_2022_CVPR} create an implicit morphable head avatar (I AM Avatar), where learned blendshapes and skinning fields are used to morph canonical geometry and texture fields during animation and rendering. Its non-rigid ray marching process can take up to 2 GPU days. HAvatar~\cite{HAvatar2023Zhao} uses a separate canonical and posed appearance volume in addition to a parametric conditioned NeRF representation, significantly improving disentanglement capabilities for more stable animation and rendering quality. Their method is very robust and synthesizes high-fidelity avatars, but training time also takes days. Lately, Bai et al.~\cite{Bai_2023_CVPR} introduced a high-fidelity volumetric head avatar solution by incorporating a 3DMM-anchored neural radiance field, which is decoded from local features using convolutional neural networks. While superior fidelity was demonstrated, both training and rendering are prolonged.

Gao et al. ~\cite{Gao_2022} presented a highly efficient video-based 3D head modeling method for NeRF rendering. They combine multi-level voxel fields with expression coefficients in latent space to reduce the neural model optimization to only 10-20 minutes. Zielonka et al.~\cite{INSTA2023Zielonka} suggest a fast reduced the required time to less than 10 minutes while outperforming the rendering quality of several state-of-the-art methods. They employed a surface-embedded dynamic NeRF approach based on multi-resolution hash encoding for instant neural graphics primitives~\cite{mueller2022instant} (Instant-NGP), combined with a 3DMM-driven geometry regularization for improved pose extrapolation.

In practice, covering all possible facial expressions during the short capture span is difficult, especially for an untrained end-user. To this end, Cao et al.~\cite{CaoChen2022authentic} introduced an approach that customizes a universal prior model trained from 255 subjects to a specific person's face. The recording session consists only of a face scan using a handheld video camera, and additional expression scans can be optionally provided to improve expressions' faithfulness.

Zheng et al.~\cite{zheng2023pointavatar} recently introduced a hybrid deformable point-based representation that can better represent intricate structures, e.g., fluffy hair, and is almost a magnitude faster to train (e.g., 6 hours instead of 54 hours). Furthermore, the neural avatars are relightable. An even faster relightable neural avatar framework (15 minutes of optimization) has been presented in ~\cite{bharadwaj2023flare}, which uses a mesh-based representation and a combination of traditional graphics rendering and neural network that approximates some of their components.

Since many of these techniques rely on a 3DMM model to capture a driver's expressions, NeRF-based cross-identity reenactment can deteriorate when there is a misalignment between the tracking results. Consequently, Xu et al.~\cite{Xu2023LatentAvatar} proposed to learn latent expression codes as the driver signal to improve the expressiveness of neural avatar heads. The method is self-supervised using a photometric reconstruction loss. To reproduce fine-scale details in facial expressions, Chen et al.~\cite{Chen_2023_CVPR} use part-based implicit shape models to decompose a global deformation field into local ones. While high-resolution details around eyes and teeth can be obtained, the method struggles with extreme expressions and poses. While significantly easier to deploy in a consumer setting, monocular video-based avatar modeling techniques still require a few minutes of facial recording. To cover all the necessary ranges of facial motion, visual guidance is often preferred, but may be hard to follow without practice. For instance, Apple's Persona avatars, released with the Apple Vision Pro, take around a minute to guide the user through various expressions and head rotation by asking the user to point the front of the VR headset to their face.

\paragraph{2D One-Shot Neural Head Reenactment.}

One-shot head reenactment techniques use only a single source image as input and a driver video to animate the source portrait. 
Since only one facial expression is available from the source (not necessarily neutral), these methods are typically based on deep generative models trained on large in-the-wild face datasets so that unseen and plausible facial expressions of the source can be generated. Moreover, as the source portrait is generally an unconstrained image, its identity needs to be disentangled effectively from arbitrary poses and expressions.

Due to the large availability of unconstrained large face image/video datasets, a number of 2D-based one-shot neural head reenactment techniques have been introduced~\cite{Wiles18,Siarohin_2019_CVPR,siarohin2019first,zakharov2019few,burkov2020neural,zakharov2020fast,doukas2020headgan,song2021pareidolia,wang2021latent,wang2021one,Ren_2021_ICCV,siarohin2021motion,EAMM_2022,hong2022depth,Tao_2022_CVPR,StyleHEAT_2022,drobyshev2022megaportraits,Zhao_2022_CVPR,Zhang_2023_CVPR,Gao_2023_CVPR,xiang2020one} and to accommodate personalized facial expressions, few shot methods have also been developed~\cite{zhang2022fdnerf,rochow2024fsrt}.

Several facial modeling techniques have been introduced, such as based on facial keypoint or 3DMMs, to separate identity, expression, and poses from the source and driver input. Despite its robustness, using linear face models to capture expressions, often limits the range of motions and expressiveness, and leads to identity leakage from the source subject during reenactment. Want et al.~\cite{wang2022progressive} proposed an implicit disentanglement approach for controllable talking head synthesis by leaning separate encoders for head pose, mouth and eye movements, and emotional expressions, where lip motions are learned from audio-visual data. Drobyshev et al.~\cite{drobyshev2022megaportraits} introduced MegaPortraits, a robust and high-resolution real-time technique based on a CNN architecture, where signals from the driver are directly mapped as facial expression features for face disentanglement and face synthesis, without using an explicit parametric model. The method was able to achieve state-of-the-art cross-reenactment results and image resolution using a carefully designed training process. To further enhance the facial expressiveness, a new model EMOPortraits~\cite{drobyshev2024emoportraits} uses a combination of latent facial expression space development and novel self-supervision loss functions, which mitigates identity information in latent expression vectors explicitly.

Despite their ability to generate impressive high-resolution facial cross reenactment results, 2D-based real-time one-shot cross-reenactment methods cannot ensure identity or expression consistency for novel views which is an important requirement for 3D rendering applications such as immersive telepresence. Hence, they are typically limited to 2D video manipulation use cases. While recent advancements in diffusion models~\cite{ho2020denoising,DBLP:journals/corr/abs-2011-13456,song2022denoising} are showing highly diverse image generation capabilities and learning capacities, several conditional diffusion-based 2D face reenactment methods have emerged~\cite{oh2024keypoint, tao2023learning, bounareli2024diffusionact, xie2024xportrait}. Most noteworthy, is the X-Portrait~\cite{xie2024xportrait} framework, which learns a motion control module directly from the driving video using a hierarchical motion attention approach. While being a purely 2D method, the novel view synthesis is highly view-consistent, but, like most diffusion-based approaches, the performance is far from real-time and the generation of realistic faces often appears cartoonish.
%
%
%
%
\paragraph{3D One-Shot Neural Head Reenactment.}

Lately, a number of 3D-aware one-shot neural head reenactment techniques~\cite{Schwarz2020graf,Chan_2021_CVPR,Niemeyer_2021_CVPR,chan2022efficient,Or-El_2022_CVPR,epigraf2022,Xue_2022_CVPR,Deng_2022_CVPR,Xiang_2023_ICCV,An_2023_CVPR,xu2022pv3d,Xu_2023_CVPR,li2023generalizable,tran2023voodoo, ki2024learning, ye2024real3dportrait} have been introduced, which aim at generating consistent renderings from novel camera poses and extreme angles, as a motivation to improve facial disentanglement and enable multi-view rendering capabilities for 3D holographic displays~\cite{li2023generalizable,stengel2023,tran2023voodoo}.  
Some of the early methods such as ROME~\cite{Khakhulin2022ROME} use an explicit mesh-based representation, e.g., FLAME~\cite{li2017learning}, combined with neural textures, but to ensure real-time performance, low polygonal meshes prevent these systems to produce high-resolution geometric and appearance details. Most of the 3D head reenactment methods use a radiance field-based representation due to its ability to learn and render complex appearances and intricate structures, while ensuring 3D awareness, meaning view-consistent geometries. 

The two methods, HeadNeRF~\cite{hong2022headnerf} and MofaNeRF~\cite{Zhuang_2022_ECCV}, use an implicit NeRF-based representation to control the head pose, but their compact latent vector representation does not always ensure the source subject's identity to be preserved, and intensive test-time optimization is often necessary. 
Most of the latest 3D-aware real-time reenactment techniques~\cite{Li_2023_CVPR,yu_2023_nofa,li2023generalizable, tran2023voodoo, ki2024learning, ye2024real3dportrait} implement tri-plane based representations~\cite{Chen2022ECCV,chan2022efficient} to generate neural radiance field from extracted input signal features, since they are particularly efficient computation-wise and also compatible with CNN architectures.

While parametric 3DMM models are typically used to extract facial expressions from the driver video and separate its features from the subject's identity~\cite{Khakhulin2022ROME,hong2022headnerf,ma2023otavatar,Li_2023_CVPR,li2023generalizable,yu_2023_nofa, ki2024learning, ye2024real3dportrait}, only a recently developed method called  VOODOO 3D~\cite{tran2023voodoo} can disentangle facial performances fully volumetrically and without an explicit mesh model. Similar to the work of~\cite{li2023generalizable} (GOHA), visual transformer networks were used to extract identity features from the source and expression features from the driver, but VOODOO 3D, introduces a particularly effective volumetric disentanglement technique by performing a 3D lifted frontalization~\cite{trevithick2023real} of both source and driver signals into a canonical neural radiance field. As demonstrated by~\cite{tran2023voodoo}, the approach can handle extremely challenging unconstrained source images and more plausible and natural expressions can be achieved, when compared to mesh-based disentanglement approaches such as~\cite{li2023generalizable}. 

In a new work, Portrait4D, Deng et al.~\cite{deng2024portrait4d} have shown that volumetric disentanglement is possible without 3D lifted frontalization for the driver, but can be achieved by directly mapping its expression features into transformer blocks of the source's 3D lifting module. The facial expressions are then neutralized by several transformer blocks, then added back at the end. To facilitate disentanglement in this 3D-aware reenactment model, the authors use a pre-trained expression encoder~\cite{wang2022progressive} trained with audio-visual data (dialogues). Shortly after its release, an improved version, Portrait4D-v2~\cite{deng2024portrait4dv2} was presented, which adopts a similar training approach as Voodoo 3D~\cite{tran2023voodoo}, where 3D lifting is first trained then the expression module, rather than both together. Compared to the first version, the results are significantly improved, producing less artifacts and handling more challenging source images. Since the pre-trained expression encoder~\cite{wang2022progressive} only disentangles control over lip motion, eye movements, and emotional expressions, very limited asymmetric facial movements are present nor intense expressions or tongues.

Our proposed work, VOODOO XP, also maps expression features of the driver directly into transformer blocks of the 3D lifted source frontalization, but does not use a pre-trained expression encoder, nor 
a neutralization step. Instead we learn the expressions encoder end-to-end use a multi-phase training approach, which combines a coarse-to-fine training approach with 
explicit neutralization and 3D lifted frontalization in its initial stage, and GAN-based fine tuning using unconstrained large video datasets. While~\cite{tran2023voodoo} have demonstrated their reenactment solution on 3D holographic displays, we presents here, the first immersive telepresence system with VR headsets, that adopts a one-shot head reenactment technique for modeling and driving 3D avatars.

\begin{figure*}[ht]
    \centering
    \includegraphics[width=0.9\linewidth]{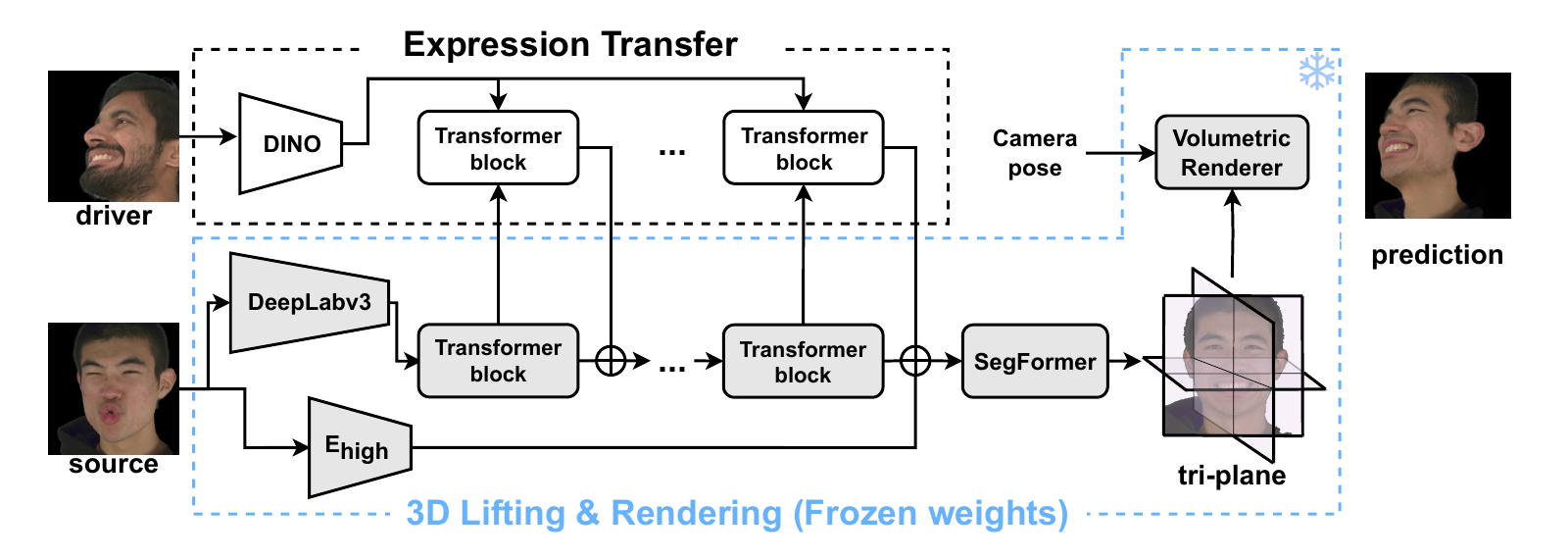}
    \vspace{-0.2cm}
    \caption{Architecture of our expressive 3D-aware head reenactment pipeline. First, a vision transformer module, DINO, is used to extract the expression vector from the driver image. This vector is then used to directly modify intermediate low-level features within the 3D lifting module using several cross and self-attention layers, altering the expression of the tri-plane representation of the source image. Finally, we can render the reenacted source tri-plane from an arbitrary viewpoint using a volumetric renderer.}
    \label{fig:architecture}
    \vspace{-0.1cm}
\end{figure*}

\begin{figure}[ht]
    \centering
    \includegraphics[width=0.95\linewidth]{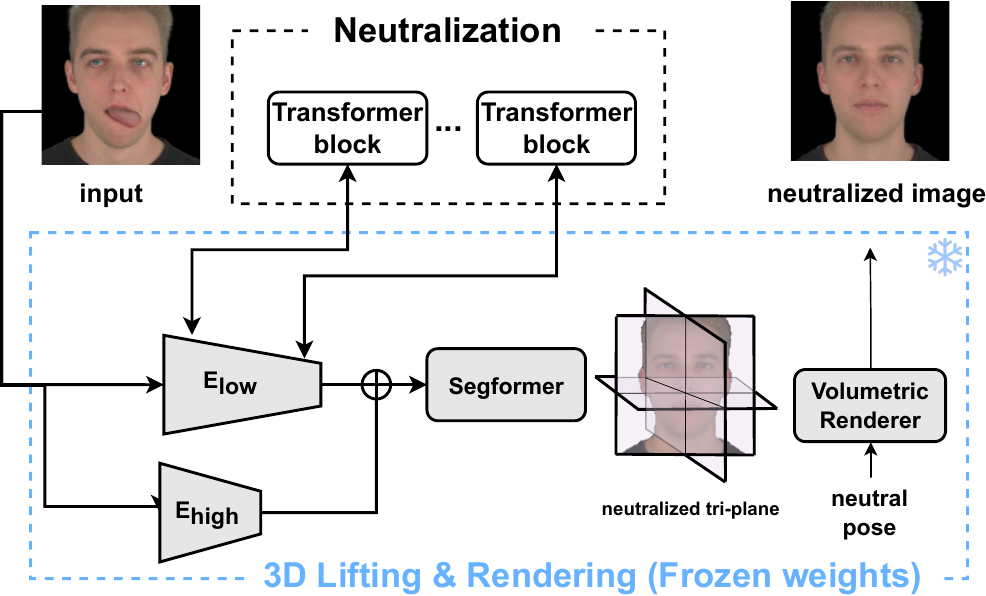}
    \vspace{-0.2cm}
    \caption{Architecture of the neutralizer. Our neutralizer is based on Lp3D but has a few additional self-attention blocks inside of the $E_{low}$ module. We only optimize these additional transformer blocks during training.}
    \vspace{-0.1cm}
    \label{fig:neutralizer}
\end{figure}

\section{Expressive 3D-Aware Head Reenactment}
Given a source image $x_{s}$ and a driver image $x_{d}$, we develop an expressive 3D-aware head reenactment model to synthesize a 3D head that combines the appearance of $x_{s}$ and the expression of $x_{d}$, allowing free-view rendering. We adopt a tri-plane based NeRF ~\cite{chan2022efficient} as the underlying 3D representation for its fidelity, generalizability, and efficiency. Our model is trained using a three-stage approach, where all the stages share the same architecture as illustrated in \cref{fig:architecture}. Specifically, on top of a pre-trained 3D lifting network \cite{trevithick2023real} (Section~\ref{sec_lp3d}), we add a trainable expression transfer 
module (Section~\ref{sec_exp}) to alter the expression. This module first uses a vision transformer initialized from DINO \cite{caron2021emerging} to extract an expression vector from the driver image $x_{d}$. 
The extracted expression vector is then used to directly modify intermediate low-level features within the 3D lifting module through several cross and self-attention layers, altering the expression of the source image $x_{s}$. However, training such network is non-trivial as shown by ~\cite{deng2024portrait4d,deng2024portrait4dv2}, where they use a pre-trained expression encoder~\cite{wang2022progressive} to facilitate disentanglement. However, their approach cannot produce expressive expressions since it is trained mainly with audio-visual data.

To stabilize the self-supervision of our transformer-based expression transfer model, we divide the training into multiple stages (Section~\ref{sec_trainig}). The first stage uses several techniques to prevent identity leakage, so that the identity information of the driver does not leak into the reenacted image. These methods include driver 3D lifted head frontalization, augmentation, and a novel neutralizing loss to minimize the difference between the neutralized source and the neutralized cross-reenacted image. In the second stage, we introduce a fine-scale training strategy using the model from the first stage to supervise a higher resolution model training with the help of generated synthetic drivers. This approach allows us to disable all constraints used in the previous step that might compromise expression quality, thereby further enhancing the expressiveness of the output. In the third stage, we adopt global fine-tuning, by unfreezing the 3D lifting module and applying GAN loss to increase the high-frequency details of the reenacted images. Additionally, per-subject fine-tuning can be introduced as an optional step to improve the likeness and the expressiveness of the target identity. Note that unless specified explicitly, no fine-tuning is applied in any of the results in this paper.

\subsection{3D Lifting and Rendering\label{sec_lp3d}}
Following previous works \cite{tran2023voodoo,bai2024real}, we adopt the state-of-the-art Lp3D \cite{trevithick2023real} as our 3D lifting model, which converts a 2D input image into a 3D neural radiance field. Its architecture is given in the bottom part of \cref{fig:architecture}. Given an input portrait, Lp3D first extracts low-level features using a hybrid transformer model and high-frequency details using a convolution-based network. These two features are then fused using several computationally efficient transformer blocks \cite{xie2021segformer} to produce the canonicalized tri-plane representation $\{T_i \in \mathbb{R}^{H \times W \times C} | 0 \le i \le 2\}$ of the input image, which can be then rendered at arbitrary viewpoints using volumetric rendering \cite{mildenhall2021nerf} with density and color. These output values can be calculated from the tri-plane as following:
\begin{align}
    c_{xyz}, \sigma_{xyz} = D(T_0(x, y), T_1(y, z), T_2(z, x)),
\end{align}
where $T_i(u, v)$ is a feature vector calculated by interpolating $T_i$ at position $(u, v)$, and $D$ is an MLP-based decoder that converts tri-plane features into color and density at each specified 3D location. To improve the rendering efficiency, Lp3D renders at a low resolution ($128 \times 128$) and then upsamples it to a higher resolution image ($512 \times 512$) using a 2D super-resolution network. We train this module using a combination of synthetic multi-view images generated by a 3D-aware face generative model EG3D~\cite{chan2022efficient} and DiffPortrait3D~\cite{gu2023diffportrait3d}, as well as a real-world multi-view dataset, NeRSemble~\cite{Kirschstein_2023}, which covers a wide range of dynamic expressions. 
%
The EG3D-based dataset is crucial for the training process since it allows us to effectively invert its internal 3D representation by minimizing the differences between the predicted and ground-truth tri-planes, as was proposed in prior work~\cite{trevithick2023real}.
%
The NeRSemble dataset has a high diversity of expressions, yet is limited in terms of identities, containing around 200 subjects.
Thus, we further boost our training data using a state-of-the-art novel-view synthesis method DiffPortrait3D~\cite{gu2023diffportrait3d}.
We use it to generate novel views for the collection of in-the-wild images with diverse appearances and expressions.

\subsection{Expression Transfer\label{sec_exp}}
As shown in previous works \cite{trevithick2023real,bai2024real}, the first branch of Lp3D extracts global features, such as the geometry of the head, while the second branch extracts high-frequency details. Therefore, given a pre-trained Lp3D model, we can modify the expression of a given source image by directly modifying its intermediate features in the low-level branch. To this end, we propose a new architecture for expression transfer as shown in the upper part of \cref{fig:architecture}. First, we use a powerful vision transformer encoder $E_{exp}$ \cite{Dosovitskiy2020AnII} initialized from a pre-trained self-supervised model \cite{caron2021emerging} to extract an expression vector $e_d = E_{exp}(x_d)$ from the driver. Then for each transformer block $B_i^{low}$ in the low-level branch of Lp3D, we use another transformer block $B_i^{exp}$ to modify its output:
\begin{align*}
    F_i = B_i^{low}(F_{i - 1}) + B_i^{exp}(F_{i - 1}, e_d),
\end{align*}
where $F_i$ is the output of $i^{th}$ transformer block in Lp3D. $B_i^{exp}(F_{i - 1}, e_d)$ uses a self-attention layers applied on $F_{i-1}$ following by a cross-attention layer with $e_d$. Different from previous works \cite{li2023generalizable,chu2024gpavatar,Khakhulin2022ROME,ma2023otavatar}, which rely on predicted expression coefficients of morphable models~\cite{10.1145/311535.311556,li2017learning} or auxiliary signals such as facial key points or audio, we directly use the RGB image of the driver to estimate the expression. This approach enables the modeling of complex expressions, such as asymmetric expressions, face squeezing, or tongue movements. However, training such an expression module is hard due to identity leakage from the source. Specifically, without proper training strategy, the identity clues of the driver, such as hair, head geometry, or skin color, can leak into the reenacted image. To address this issue, we introduce an efficient three-stage training strategy, as detailed in the next section.

\subsection{Multi-stage Training\label{sec_trainig}}
\paragraph{Stage 1: Coarse-Level Model.} To train the expression transfer module, we propose a three-stage approach. In the first stage, we follow the standard training pipeline of other facial reenactment methods. Given a large-scale facial video dataset, during training, we randomly sample two frames $x_{s_1}$ and $x_{d_1}$ from the same video and use the following loss to train the model:
\begin{align*}
    \mathcal{L}_{rec}(x_{s_1 \rightarrow d_1}, x_{d_1}) = \mathcal{L}_{1}(x_{s_1 \rightarrow d_1}, x_{d_1}) + \mathcal{L}_{per}(x_{s_1 \rightarrow d_1}, x_{d_1})\\ + \mathcal{L}_{ID}(x_{s_1 \rightarrow d_1}, x_{s_1}),
\end{align*}
where $x_{s_1 \rightarrow d_1}$ is the reenacted image with the source image $x_{s_1}$ and the driver image $x_{d_1}$, $\mathcal{L}_{1}$ the L1 distance between the reenacted image and the the driver image, $\mathcal{L}_{per}$ the perceptual loss \cite{zhang2018perceptual} between them,
and $\mathcal{L}_{ID}$ measures the negative cosine similarity between the face recognition embeddings \cite{deng2019arcface} of the facial images. Since the model is trained exclusively on self-reenactment data, it can transfer not only the expression but also some identity clues from the driver to the reenacted image. This phenomenon is known as identity leakage. To avoid this, we adopt several techniques from previous works, including frontalizing the driver using a 3D lifting module \cite{tran2023voodoo} and applying strong augmentations to the frontalized driver image \cite{tran2023voodoo,drobyshev2022megaportraits}, such as random color patching and random masking.
However, these techniques alone cannot completely solve the identity leakage problem. To fully address this issue, we propose a novel neutralizing loss. Specifically, let $\mathcal{N}(x)$ be a function that maps the input face image $x$ with arbitrary expression and pose into its neutralized tri-plane representation. We then add the following neutralizing loss to ensure the identity matching between the source image and the cross-driver reenacted image:
\begin{align*}
    \mathcal{L}_{neu}(x_{s_1}, x_{d_2}) = \|\mathcal{N}(x_{s_1 \rightarrow d_2}) - \mathcal{N}(x_{s_1})\|
\end{align*}
where $s_1$ and $d_2$ are two identities from different videos. Finding $\mathcal{N}$ is a highly challenging task due to the lack of supervised data for training. Inspired by Noise2Noise \cite{pmlr-v80-lehtinen18a}, which demonstrated that denoising networks can be trained on pairs of noisy images to predict the underlying clean signal by averaging out the noise without the need of seeing the ground truth clean image, we consider the following optimization task:
\begin{align}
    \mathcal{N}^* = \argmin_{\mathcal{N}} \mathbb{E}_{x_{s_1}, x_{d_1}}\|\mathcal{N}(x_{s_1}) - T_{d_1} \|, \label{eq:neu}
\end{align}
where $x_{s_1}$ and $x_{d_1}$ are statistically independent given that they share the same identity, and $T_{d_1}$ is the tri-plane representation of $x_{d_1}$, which can be extracted by the pre-trained 3D lifting module. Suppose we have enough data, then the optimal function $\mathcal{N}^*$ will converge to the average value of all $T_{d_1}$ given the identity of $x_{s_1}$, representing a tri-plane of the neutralized face. Therefore, even without supervised neutralizing data, we can use the objective function in \cref{eq:neu} to train the neutralizer using a large-scale video dataset. We implement the neutralizer using the network structure described in \cref{fig:neutralizer}. Our final loss for the first stage can be formulated as:
\begin{align*}
    \mathcal{L}_{\text{stage 1}}(x_{s_1}, x_{d_1}, x_{d_2}) =  \mathcal{L}_{rec}(x_{s_1 \rightarrow d_1}, x_{d_1}) + \lambda_{neu}\mathcal{L}_{neu}(x_{s_1}, x_{d_2}),
\end{align*}
where $\lambda_{neu}$ is the weight for the neutralizing loss. In our experiment, we set $\lambda_{neu} = 0.01$.

\paragraph{Stage 2: Fine-Level Model.} 

While the model trained in the first stage can produce reasonable expressions without identity leakage, the techniques used in that stage, such as driver frontalization, augmentations, and neutralizing loss, can potentially clamp the expressiveness. In addition, for better training efficiency, the resolution used in the first stage is only $128 \times 128$. Previous works \cite{deng2024portrait4d,deng2024portrait4dv2,tran2023voodoo,li2023generalizable} uses a $4\times$ super-resolution network to upsample the output to $512 \times 512$. However, this approach often leads to inconsistencies, especially with extreme poses, as shown in \cref{fig:qual}. Therefore, we introduce a second stage to improve both the expressiveness and resolution of the first stage.

We first increase the rendering resolution of the 3D lifting module from 128 to 256. We initialize the new 3D lifting module with the weights from the first step, then fine-tune it at the higher rendering resolution using the same set of losses. During training, instead of calculating losses on the $256 \times 256$ images, we randomly render patches of sizes $172 \times 172$ in order to avoid the GPU memory limitations and facilitating the use of larger batch sizes.

For the expression transfer module, we use the same architecture of the first stage and initialize the network with the weights of the trained model. To further improve expressiveness, we remove driver frontalization, data augmentations, and the neutralizing loss. To overcome the identity leakage problem, resulting from removing these techniques, and instead of only using paired source and driver as self-reenactment data in the first stage, we use the first-stage network to synthesize a cross-reenactment dataset. For each synthetic data point, the source $x_{s_1}$ and the ground-truth $x_{d_1}$ are sampled from the same video. Additionally, instead of only using $x_{d_1}$ as the driver as in self-reenactment training, we generate a synthetic driver $x^{1}_{s_2 \rightarrow d_1}$ using the reenactment network, trained in the first stage. This network uses $x_{d_1}$ to drive a different identity image $x_{s_2}$ sampled from another video. This data synthesizing approach uses real images as both the source and ground truth, while only using the output of the first stage as the driver. Consequently, the quality of the second stage is no longer constrained by the first stage. Finally, we adopt an eye-region loss $\mathcal{L}_{\text{eye}}$ on self-reenactment data, where only the eye regions are rendered using the frontal view of the reenacted image $x_{s_1 \rightarrow d_1}$ and ground-truth image $x_{d_1}$, and we calculate the $L_1$ loss between them. As can be seen in \cref{fig:eye_qual}, this loss significantly improves the eye gaze accuracy of the output. The loss for the second stage can be summarized as follows:
\begin{align*}
\mathcal{L}_{\text{stage 2}}(x_{s_1}, x_{s_2}, x_{d_1}) =  \mathcal{L}_{\text{rec}}(x_{s_1 \rightarrow x^{1}_{s_2 \rightarrow d_1}}, x_{d_1}) +  \mathcal{L}_{\text{rec}}(x_{s_1  \rightarrow d_1}, x_{d_1}) \\
+ \mathcal{L}_{\text{eye}}(x_{s_1  \rightarrow d_1}, x_{d_1}).
\end{align*}

\paragraph{Stage 3: Global fine-tuning.} The model from stage 2 can produce highly expressive faces at high-resolution. However, we notice that in practice, it does not preserve the identity of the source well. Note that this issue is not due to identity leakage, as the identities of the reenacted images remain consistent across different drivers. We conjecture that this issue arises from the following reasons: (1) We only uses $L_1$ and LPIPS losses, which cause the model to ignore high-frequency details; (2) Most of the data used in the previous stages are expressive, but limited in the number of identities~\cite{Kirschstein_2023}; and (3) In both stages, the 3D lifting module is frozen, which limits the capacity of the expression transfer module. Based on these hypotheses, we introduce a final stage where we fine-tune the entire network with an additional GAN loss. Specifically, we unfreeze the weights of the lifting module and then train the model with the following loss:
\begin{align*}
\mathcal{L}_{\text{stage 3}}(x_{s_1}, x_{s_2}, x_{d_1}) =  \mathcal{L}_{\text{rec}}(x_{s_1 \rightarrow sg(x_{s_2 \rightarrow d_1})}, x_{d_1}) \\ +
\mathcal{L}_{\text{rec}}(x_{s_1 \rightarrow x_{d_1}}, x_{d_1}) + \lambda_{GAN}\mathcal{L}_{\text{GAN}}(x_{s_1 \rightarrow sg(x_{s_2 \rightarrow d_1})}, x_{d_1})
\end{align*}
where $sg(\cdot)$ is the stop-gradient operator and $\mathcal{L}_{\text{GAN}}$ is a standard projected GAN loss \cite{sauer2021projected} with corresponding weight $\lambda_{GAN}=0.05$ in our experiment. Here, we use a similar idea as in stage 2, where we leverage the previous model to synthesize the driver and therefore avoid identity leaking. However, different from the second stage, we perform this on-the-fly, using a stop-gradient operator. We observe that this stage greatly improves the identity preservation and high-frequency expression details of the reenacted images.

Finally, we use the trained network to synthesize a set of pairs $x_s \in \mathbb{R}^{512\times512\times3}$ and $x_{s \rightarrow s} \in \mathbb{R}^{256\times256\times3}$ and use them to train a consistent $2\times$ super-resolution network \cite{wang2021real}. In \cref{fig:qual}, our results on extreme poses and novel view synthesis are much more consistent than other methods.

\paragraph{Optional step: Few-shot fine-tuning.} While our proposed work is a one-shot reenactment technique, it inherently supports few-shot input as a fine-tuning process. In particular when a few images or video of the source is available, we simply use these additional source images to fine-tune our model, using he same training strategy as in stage 3. Thanks to the loss used in stage 3, the fine-tuned model still works with an arbitrary driver. This optional stage can significantly improve the expressiveness and likeness of the reenacted result, even with a few additional images (e.g., 10 or 20), as demonstrated in the results section. In practice, this optional step takes no longer than additional tens of seconds of computation for each source subject.

\subsection{Implementation Details}
We train the models of each stage using a mix of CelebVHQ \cite{zhu2022celebv}, which is a large-scale video dataset with more than 33000 number of identties, and Nersemble \cite{Kirschstein_2023}, which contains a large variety of expressions, but the number of identities is limited (. We train the $128 \times 128$ 3D lifting module for five days on 7 NVIDIA RTX6000ADA with a batch size 14 and a learning rate of $1e-4$. For the $256 \times 256$ 3D lifting module, it only takes a day for the model to converge since we fine-tuned it from the weights of the low-res version. The expression module takes five days for all three stages to converge.

\section{VR Telepresence System}

\begin{figure}
    \centering
    \includegraphics[width=0.95\linewidth]{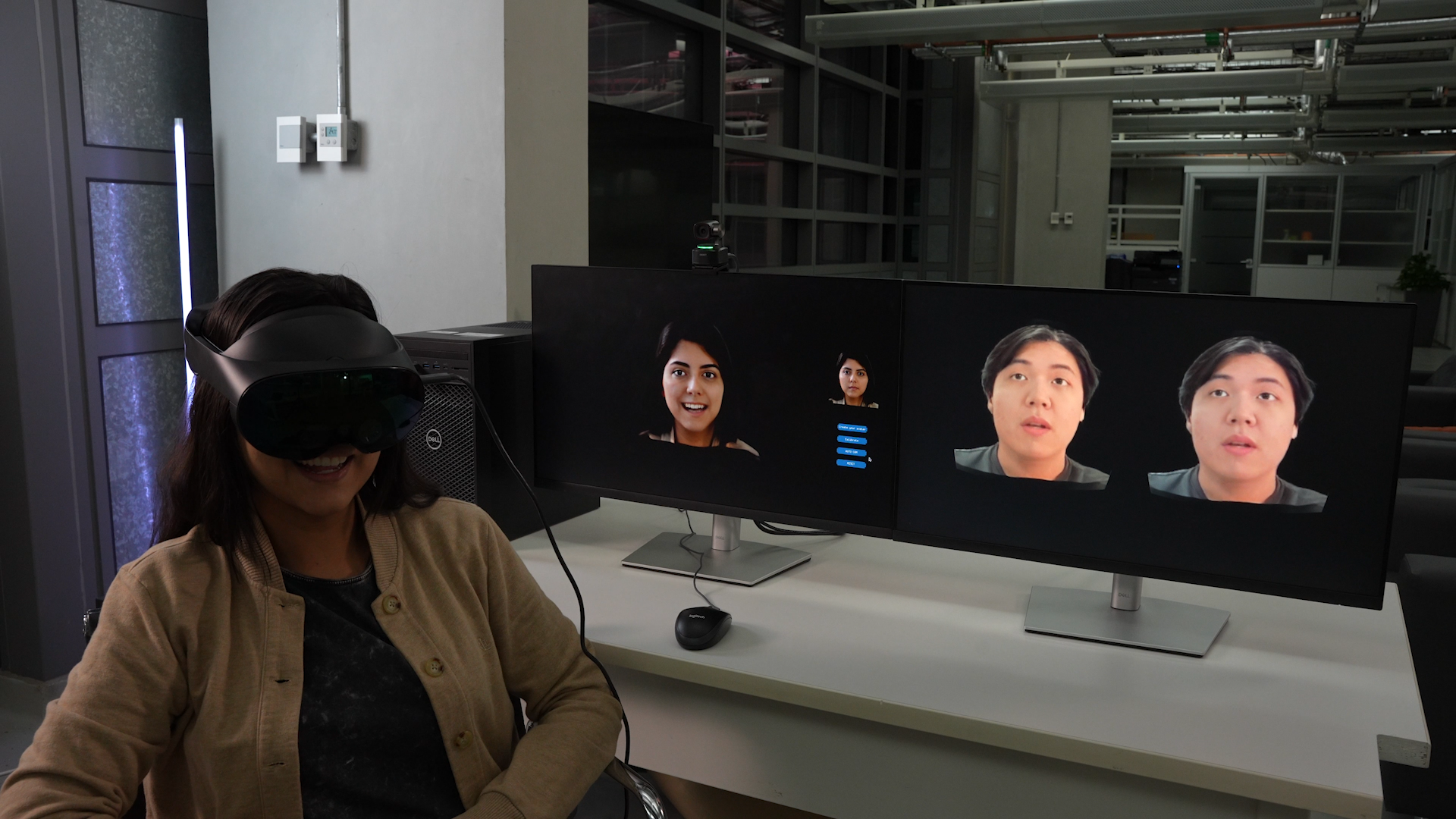}
    \vspace{-0.2cm}
    \caption{VR telepresence system setup. Our solution can be applied to an end-to-end VR telepresence system for two-way communication.}
    \label{fig:vrsystem}
\end{figure}

We propose the first immersive telepresence system for two-way remote communication based on realistic 3D avatars built from one-shot neural head reenactment. Each participant uses a Meta Quest Pro VR headset, tethered to a high-performance workstation with four NVIDIA RTX 6000 ADA GPUs (see Figure~\cref{fig:vrsystem}). In particular, 2 GPUs are used for each eye (left and right) for stereoscopic rendering and we achieve 30 FPS at $512\times 512$ pixel resolution (including super resolution from $256\times 256$). Our solution allows users to instantly build their 3D avatar instantely ($60$ ms), just by taking a photo with a webcam or by loading a picture. Once created, the users can immediately control their neural heads by simply wearing their VR headset, which captures their facial expressions and eye-gaze, using the built-in outside-in cameras and the 6DoF head pose using the integrated inside-out/IMU tracker.

\paragraph{System overview.}

\begin{figure}
    \centering
    \includegraphics[width=0.95\linewidth]{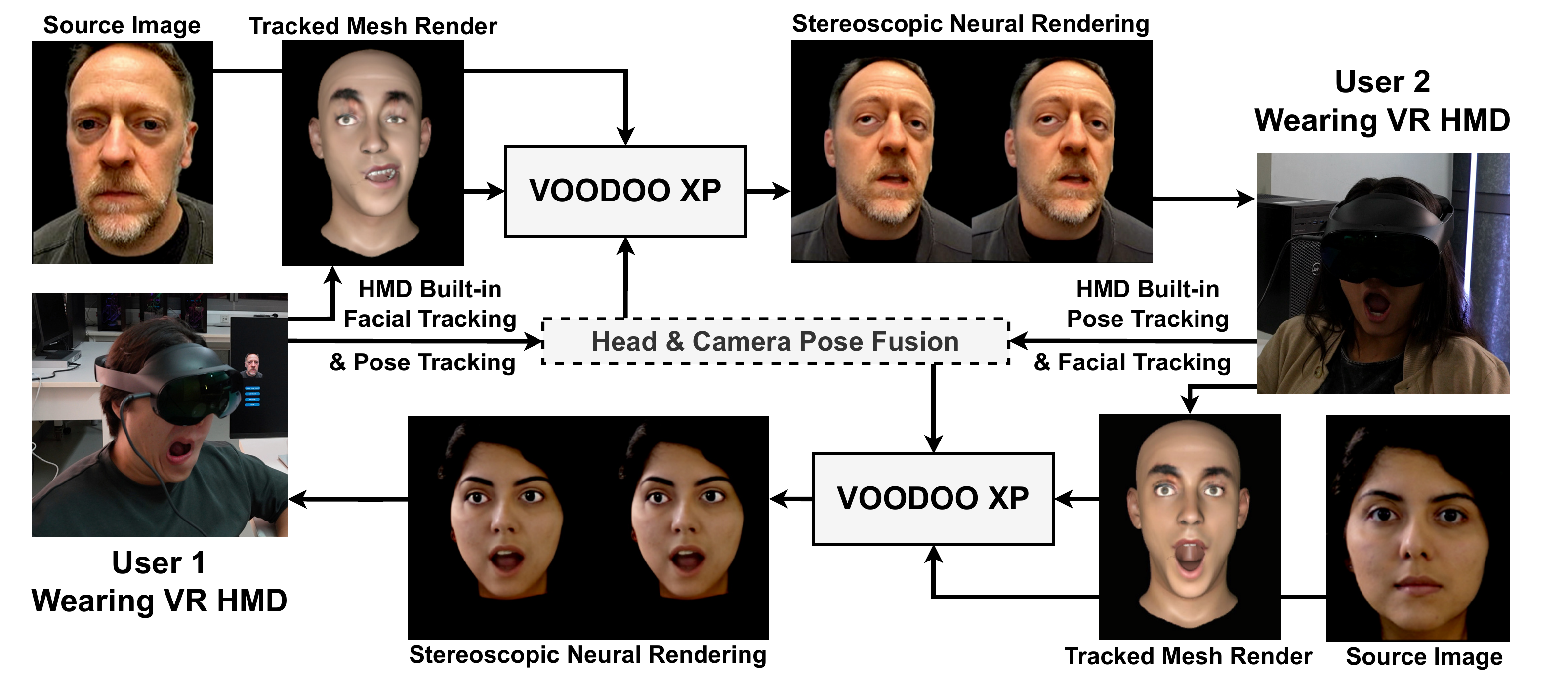}
    \vspace{-0.1cm}
    \caption{Overview of our two-way VR telepresence system. For each user, we track pose and expressions using built-in tracking of the headset and use them to drive our reenactment system. To create an avatar, we can either use a user's portrait (User 2) or an arbitrary photo (User 1).}
    \label{fig:vr_system}
    \vspace{-0.2cm}
\end{figure}

Our complete VR telepresence system is illustrated in \cref{fig:vr_system}. For VR-based facial performance capture, we simply use Meta's Movement SDK~\cite{metamovementsdk} and Headset Tracking SDK ~\cite{headsettrackingsdk}, as input signals to a Unity game engine~\cite{unityengine} scene with a generic parametric blendshape model. The facial expressions consist of 63 blendshapes (7 for tongue) and 2 gaze controls (angles) for each eye. This generic face model is then animated and rendered using a traditional computer graphics (CG) pipeline to produce a live video stream, which serves as input to our neural head reenactment framework. 

For each person's Unity scene, we render the generic face model placed in a canonical coordinate frame without applying 6DoF head pose transformations. Then, our VOODOO XP framework converts a CG video of a frontalized generic avatar into a photorealistic avatar, seen from two novel stereoscopic views. Two instances of VOODOO XP are running in parallel on the workstation and all frame processing is fully synchronized using another controller thread.
Note that user 1 sees the face of user 2, so the avatar creation, animation, and rendering on user 1's side, should be for the avatar of user 2. Hence, our system transfers user 2's avatar input photo first onto the system of user 1, and vice versa, before communication happens. While the rendering of user 2's avatar occurs on user 1's side, it uses the head pose of user 1 as the camera view pose during reenactment. Consequently, the inverse head pose of user 2 is multiplied by the head pose of user 1 on user 1's system. We denote this step as head and camera pose fusion in \cref{fig:vr_system}.
Finally, the combined head poses are used as input for novel view generation, combined with the frontalized driver video for both left and right eye cameras, into our VOODOO XP head reenactment framework. The stereoscopic neural renderings are then fed back into the same local Unity scene instance and rendered into VR HMD accordingly as synchronized stereoscopic video textures via Meta XR Simulator.

Our immersive communication is extremely low bandwidth and low latency, as we only need to transmit 63 facial blendshape values, based on Facial Action Coding System~\cite{Ekman_1978_10190} (FACS), 4 signals for eye gaze angles, and 6DoF head pose parameters between users, as all the neural rendering of the opposite user happens on each person's system locally. 

\paragraph{Generic Driver Rig.}

In contrast to our base VOODOO XP system, which can process driver input frames of arbitrary subjects in unconstrained settings and head poses, we use a generic avatar head model for our VR application, with expressions compatible with Meta's Movement SDK and use a frontalized view as it maximizes the visibility of expressions independently of the user's head pose. We adopted the shape and mesh topology of the generic head model from Pinscreen's AvatarNeo SDK~\cite{pinscreenavatar} and transferred FACS-based~\cite{Ekman_1978_10190} facial expressions from the Meta Movement SDk's Aura character. Our model consists of 13K vertices for the face, 4K vertices for the mouth interior (gum, teeth, tongue), and 3K vertices for the eyes.

To build the facial expression blendshapes for our generic head model, we first warp our generic head model to the Auro neutral face mesh to establish one-to-one correspondences between meshes with different mesh topologies using a non-rigid registration algorithm based on~\cite{li09robust}. A deformation transfer technique~\cite{10.1145/1015706.1015736} is then used to transfer facial expressions from the Aura face model back to our generic head model. We transfer all 273 facial blendshapes as well as 28 additional blendshapes of the mouth interior model (gum, teeth, and tongue) from Aura's model, and combine all of them into 63 linear blendshapes for the headset tracking. A skilled digital artist refined all facial blendshapes via 3D sculpting and real-world FACS expression references corresponding to Meta's movement SDK guideline, to ensure expressive and natural movements as not all facial expressions of Aura are ideal. This process is only done once as we are using the same generic head model for each driver.


\section{Experiments}

\subsection{Benchmark Dataset}


\begin{figure}
    \centering
    \includegraphics[width=0.95\linewidth]{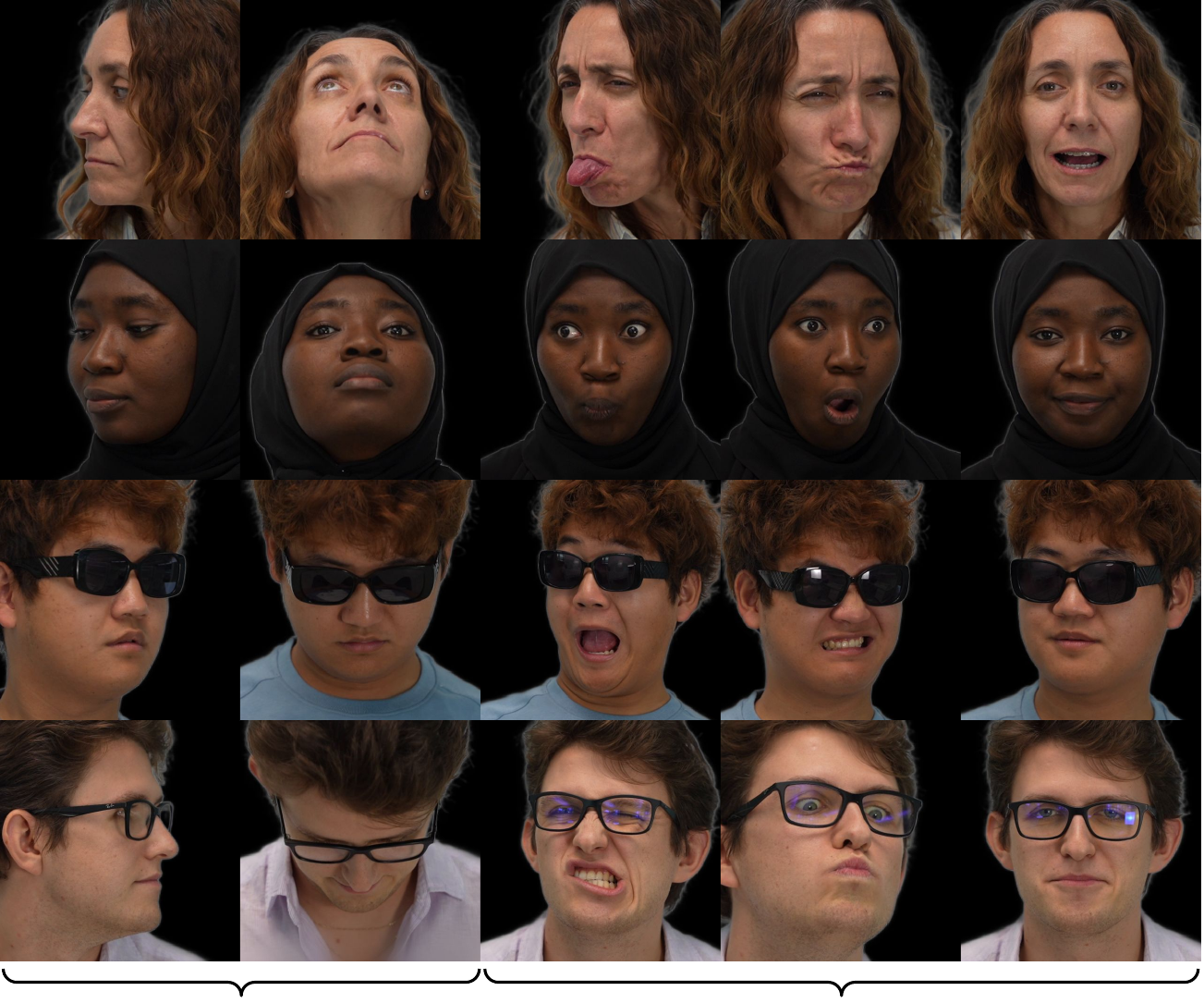}
    \newcolumntype{Y}{>{\centering\arraybackslash}X}
    \begin{tabularx}{0.95\linewidth}{YY}
         \hspace{-0.8cm} \textbf{Pose coverage} & \hspace{-0.6cm} \textbf{Expression coverage}
    \end{tabularx}
    \vspace{-0.3cm}
    \caption{Our evaluation dataset covers a wide range of poses, expressions, genders, accessories, and ethnicities}
    \label{fig:data_samples}
    \vspace{-0.1cm}
\end{figure}
\begin{figure*}
    \centering
    \includegraphics[width=0.95\linewidth]{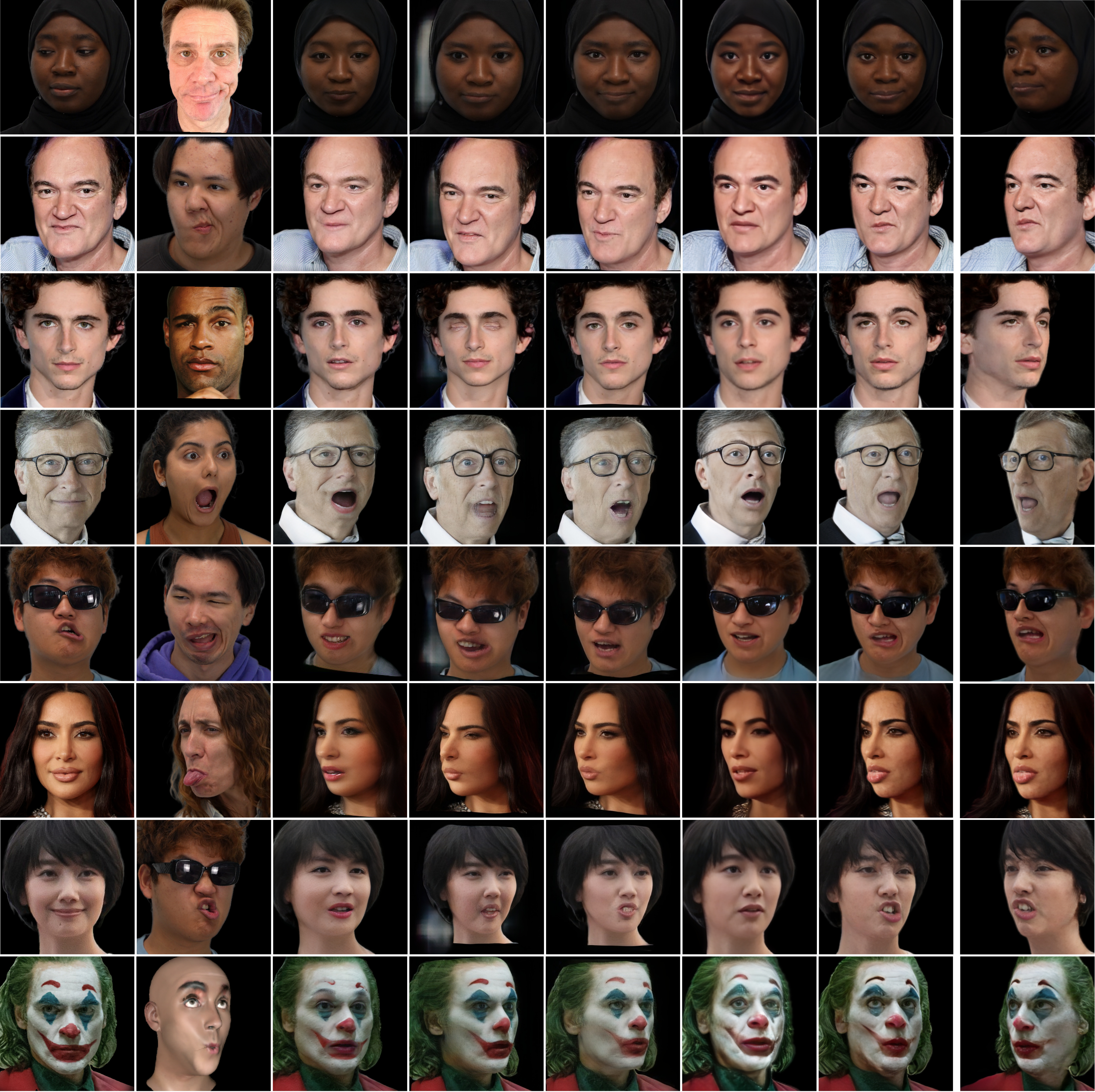}
    \newcolumntype{Y}{>{\centering\arraybackslash}X}
    \begin{tabularx}{0.95\linewidth}{YYYYYYYY}
         \textbf{Source} & \textbf{Driver} & \textbf{GOHA} & \textbf{Portrait4D} & \textbf{P4D-v2} & \textbf{VOODOO3D} & \textbf{Ours} & \textbf{(novel view)}
    \end{tabularx}
    \vspace{-0.3cm}
    \caption{Qualitative comparison in the one-shot head reenactment scenario. The last column showcases the novel view synthesis results for our method.}
    \vspace{-0.1cm}
    \label{fig:qual}
\end{figure*}
\begin{figure*}
    \centering
    \includegraphics[width=0.95\linewidth]{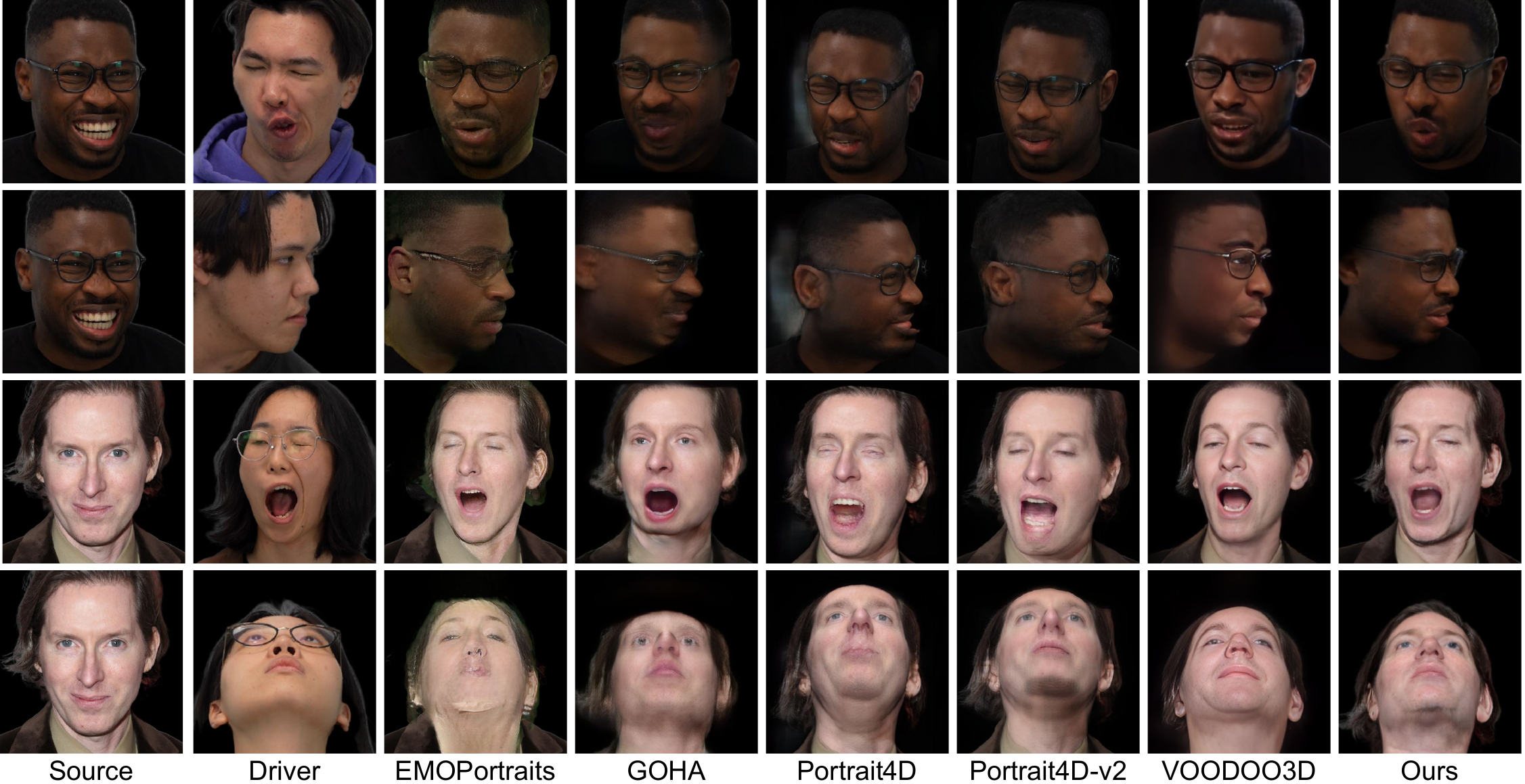}
    \newcolumntype{Y}{>{\centering\arraybackslash}X}
    \begin{tabularx}{0.95\linewidth}{YYYYYYYY}
         \textbf{Source} & \textbf{Driver} & \textbf{EMOPortraits} & \textbf{GOHA} & \textbf{Portrait4D} & \textbf{P4D-v2} & \textbf{VOODOO3D} & \textbf{Ours}
    \end{tabularx}
    \vspace{-0.3cm}
    \caption{Qualitative comparison of one-shot head reenactment given extreme pose or expression drivers. Our method can handle these corner cases consistently better than competitors.}
    \label{fig:emoportraits}
    \vspace{-0.1cm}
\end{figure*}
\begin{figure*}
    \centering
    \includegraphics[width=0.95\linewidth]{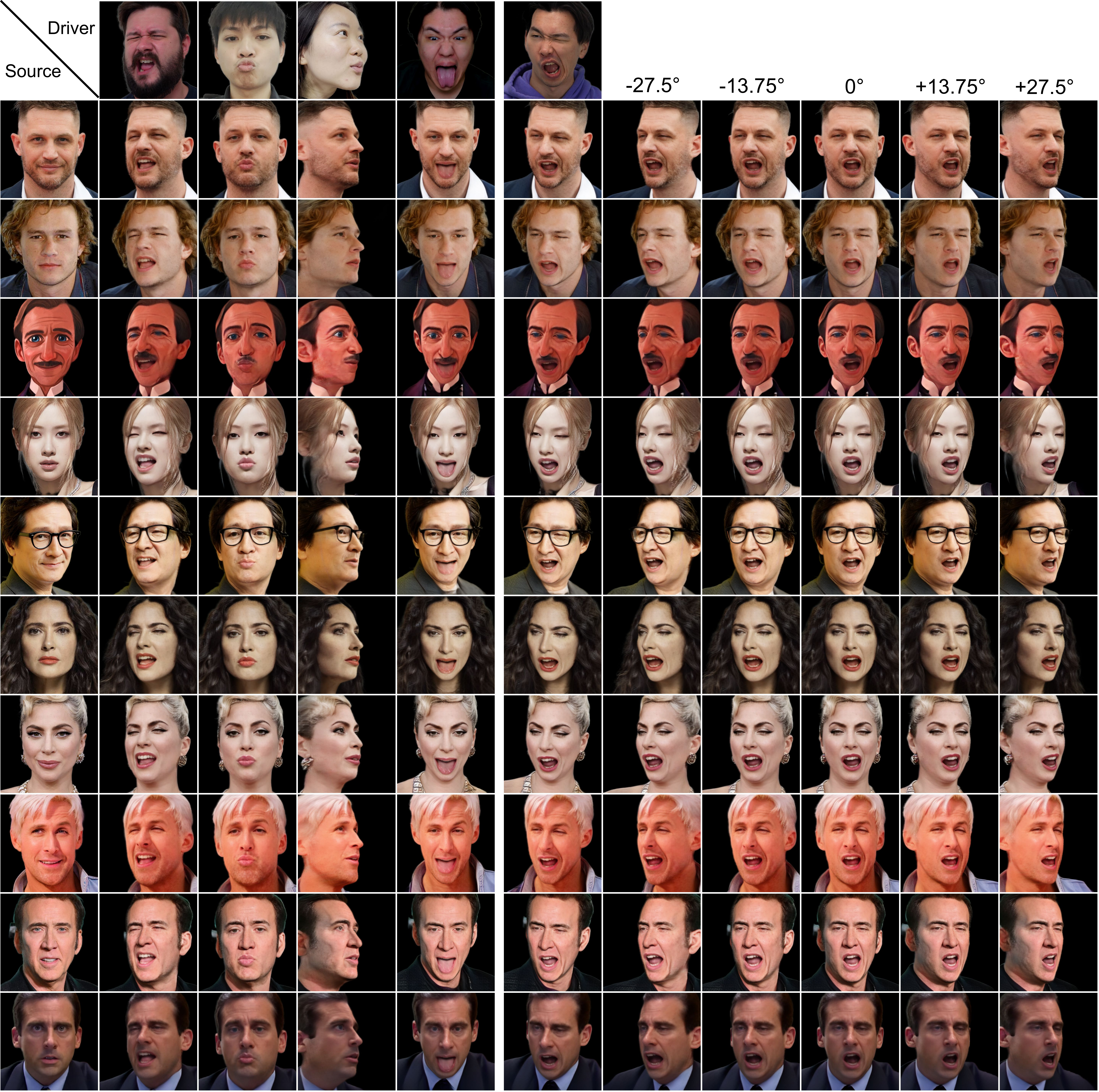}
    \vspace{-0.3cm}
    \caption{Reenactment results with different drivers show that our method consistently preserves the identity of the source and expression of the driver.}
    \vspace{-0.1cm}
    \label{fig:thematrix}
\end{figure*}
Existing datasets commonly used for benchmarking~\cite{zhang2021flow,karras2017progressive} lack pose diversity, since most faces in them are frontal. In addition, they do not cover a wide range of facial expressions. In this work, we propose a new evaluation dataset for facial reenactment that covers a wide range of different poses, expressions, and ethnicities. To this end, we capture video data of 17 different subjects. For each subject, we record five different scenarios: normal conversation, extreme pose rotation, mild pose rotation, and extreme expression with random poses. We also take into account many different accessories, e.g., hijab, sunglasses, or eyeglasses. This resulted in 102 videos totaling more than 137,000 frames. We filter out similar frames by performing a clustering of 200 centers based on each frame's facial landmarks and then pick 200 frames that have the closest landmarks to each center. This reduced the number of frames to 20,400 which we used for the evaluation.
The examples from our dataset can be seen in \cref{fig:data_samples}.

For the self-reenactment task, we use the first frame of each video as the source and the rest as the driver. For the cross-reenactment task, we also use the first frame of the videos as the source but the drivers are frames of another video with different identities. 


\subsection{Baseline Comparison}

We compare our method against the following one-shot neural head avatar systems: EMOPortraits~\cite{drobyshev2024emoportraits}, GOHA~\cite{li2023generalizable}, Portrait4D~\cite{deng2024portrait4d}, Portrait4D-v2~\cite{deng2024portrait4dv2}, and VOODOO 3D~\cite{tran2023voodoo}.
%

\textbf{EMOPortraits}~\cite{drobyshev2024emoportraits} is a 2D avatar synthesis method that models extreme expressions via latent vectors of low dimensionality.
The latter are learned in a self-supervised fashion from the in-the-wild video data.
%
%
%
%
%

\textbf{GOHA}~\cite{li2023generalizable} represents the subject's expression using a 3DMM~\cite{10.1145/311535.311556} parametric model.
Their method employs a pre-trained expression extractor~\cite{deng2019accurate} that learns to predict facial shape and texture blendshapes from in-the-wild data using photometric supervision.
These expressions are then imposed onto a tri-plane representation of the face~\cite{chan2022efficient} by predicting its residual.
%
%
%
%

\textbf{Portrait4D}~\cite{deng2024portrait4d} and \textbf{Portrait4D-v2} (denoted as \textbf{P4D-v2}) both leverage a pre-trained 3D human face generation model called GenHead that combines a modified EG3D~\cite{chan2022efficient} pipeline with explicit conditioning on the FLAME~\cite{li2017learning} head pose and expression parameters.
%
%
The authors leverage GenHead to generate a synthetic dataset used to train a 3D avatar system conditioned on a source image identity and a driver image expression.
To encode the expression from the driver image while preserving the disentanglement between the identity and expression, they rely on a pre-trained expression extractor~\cite{wang2022progressive} which is supervised using audio and keypoint-based data and thus does not contain identity-specific information.
%
%
%
%

\textbf{VOODOO 3D}~\cite{tran2023voodoo} also leverages a pre-trained generative model for human faces.
In this work, similarly to ours, the authors use the EG3D~\cite{chan2022efficient} generator to train a 3D lifting network, following the approach of Live 3D Portraits~\cite{ye2024real3dportrait}.
VOODOO 3D then uses an expression estimation procedure that modifies the 3D-lifted representation and is learned directly from in-the-wild data.
%
%
%
%
%

\subsection{Qualitative Evaluation of Cross-Reenactment}

We conduct the evaluation using both in-the-wild and studio data. For the in-the-wild data, we use a set of celebrities with diverse ethnicities, poses, and expressions.
%

\begin{figure}
    \centering
    \includegraphics[width=0.95\linewidth]{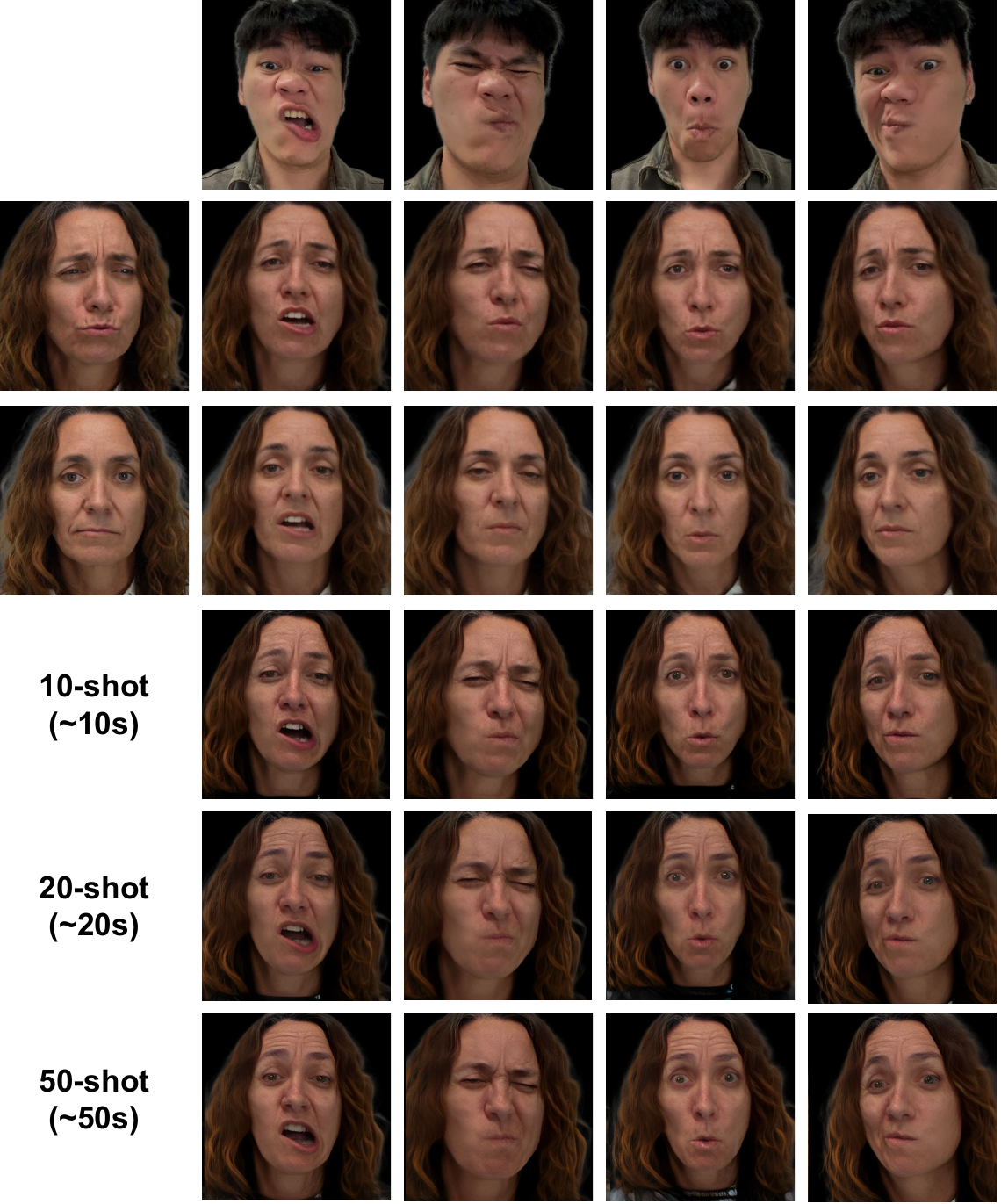}
    \vspace{-0.2cm}
    \caption{Our model can be optionally fine-tuned on a few images of the subject to significantly improve the expression. The second and third rows are one-shot reenactment that use different source images. The last three rows adopt few-shot fine-tuning which shows superior expressiveness compared to the one-shot setting.}
    \label{fig:fewshot}
    \vspace{-0.1cm}
\end{figure}

%
The qualitative comparison results are presented in \cref{fig:qual,fig:emoportraits}, more examples for our method with the novel-view synthesis evaluation in \cref{fig:thematrix}, and few-shot fine-tuning results in \cref{fig:fewshot}.
For EMOPortraits, we showcase performance only in extreme cases, as this method is 2D and is not designed for the 3D novel-view synthesis task.

First, we notice that most of the baseline methods are capable of reenacting simple expressions, such as open mouth, see the fourth row of \cref{fig:qual}.
However, both versions of Portrait4D noticeably struggle to generate images without artifacts, which are especially visible in the region of teeth, and GOHA changes the facial shape.
When even minor asymmetries are introduced in the mouth or eyebrow shapes, such as in the first three rows, it already poses a substantial challenge for the base methods.
For example, we can see that our method can faithfully convey a correct change of the mouth shape, see the example in the first row, while GOHA and VOODOO 3D simply reenact a smiling face.
None of the base methods is also capable of reenacting asymmetrically raised eyebrows, see row three.
Going to the more extreme mouth shape changes in the last four rows, we can see most of the base methods completely fail to reflect the expression of the driver, while our method manages to reenact the faces with high accuracy, especially in the mouth region. 
%
%
Lastly, even in the most challenging cases showcased in \cref{fig:emoportraits}, i.e., for large head rotations and tongue movement, our method still produces convincing results.
%
%
Notice that while VOODOO 3D is also capable of realistically synthesizing large head pose movements, it introduces substantial identity change in the process.
It is also worth noting that our method can synthesize tongues as well as correct eyebrow movements behind glass frames, as well as reconstruct view-consistent eyeglasses.
%
%

We additionally evaluate the disentanglement between identity, expression, and head pose in our method, the results are in \cref{fig:thematrix}.
In the left part of the figure, we evaluate the consistency of the identity given changes in the head pose and the expression.
On the right, we showcase the consistency of the expressions given the change in the head pose.
Our method manages to preserve the identity given both expressions and pose changes, which is an essential property for telepresence systems.
%
Lastly, we showcase few-shot fine-tuning capabilities.
Given a collection of photos of a short video of a subject, we can fine-tune our model to improve both identity preservation and the expressiveness of our model.
For the fine-tuning results, please refer to \cref{fig:fewshot}.





\subsection{Quantitative Evaluation}

\begin{table*}
    \centering
    \setlength{\tabcolsep}{3pt}
    \begin{tabular}{l|cccccccc|cccc|}
         \multirow{2}{*}{Method} & \multicolumn{8}{c|}{Self-reenactment} & \multicolumn{4}{c|}{Cross-reenactment} \\
         & PSNR $\uparrow$ & LPIPS $\downarrow$ & SSIM $\uparrow$ & CSIM $\uparrow$ & FID $\downarrow$ & AKD $\downarrow$ & AED $\downarrow$ & APD $\downarrow$ & CSIM $\uparrow$ & FID $\downarrow$ & AED $\downarrow$ & APD $\downarrow$ \\
         \hline
         GOHA \cite{li2023generalizable} &  22.21 & \textbf{0.18} & 0.69 & 0.67 & \textbf{32.20} & 7.75 & 4.28 & 0.060 & 0.63 & \underline{40.68} & 6.87 & 0.096 \\
         Portrait4D \cite{deng2024portrait4d} & 20.11 & 0.21 & 0.64 & 0.73 & 51.06 & 7.03 & 3.97 & 0.069 & 0.66 & 51.24 & 6.80 & 0.098 \\
         Portrait4D-v2 \cite{deng2024portrait4dv2} & 22.20 & 0.20 & \underline{0.70} & \textbf{0.77} & 51.23 & \underline{6.51} & \underline{3.65} & 0.063 & \textbf{0.67} & 50.14 & \underline{6.53} & 0.099 \\
         VOODOO 3D \cite{tran2023voodoo} & \textbf{23.45} & 0.21 & \textbf{0.73} & 0.69 & 48.23 & \textbf{6.36} & 4.32 & \underline{0.057} & 0.58 & 50.30 & 6.70 & \textbf{0.088} \\
         Ours & \underline{22.30} & \textbf{0.18} & 0.68 & \underline{0.75} & \textbf{23.80} & 7.29 & \textbf{3.41} & \textbf{0.050} & \textbf{0.67} & \textbf{25.80} & \textbf{6.31} & \underline{0.090} \\
    \end{tabular}
    \caption{Quantitative evaluation of our method in self- and cross-reenactment settings.}
    \label{tab:quant}
    \vspace{-0.6cm}
\end{table*}

We measure the quality of our method quantitatively in the self- and cross-reenactment scenarios.
For self-reenactment, we evaluate standard metrics that measure pixel-wise similarity, such as PSNR, LPIPS~\cite{zhang2018perceptual}, and SSIM.
We additionally measure cosine similarity between the embeddings of the face recognition network~\cite{deng2019arcface} (CSIM), which is used to measure identity preservation~\cite{zakharov2019few}; Frechet Inception Distance~\cite{10.5555/3295222.3295408} (FID), which evaluates visual image quality; and average keypoint, expression and pose distance (AKD, AED and APD), which measure the difference between the estimated 2D keypoints~\cite{bulat2017far}, expression vectors~\cite{danvevcek2022emoca} and head pose~\cite{feng2021learning}.
%
%
For cross-reenactment, we only measure the metrics that do not require pixel-wise alignment between the prediction and ground truth, namely CSIM, FID, AED, and APD.

We provide the results in Table~\ref{tab:quant}. 
Our method achieves higher image quality among the competitors in self-reenactment, achieving both the best FID and LPIPS scores.
It is also highly accurate w.r.t. the pose, achieving the best AED and APD.
We note that average keypoints distance, AKD, is highly influenced by the quality of the keypoint extractor that is commonly used to calculate this metric, which is known to fail for asymmetric and extreme expressions.
%
%
Our method is also highly competitive w.r.t. PSNR and SSIM, albeit falling short of VOODOO 3D which predicts oversmoothed results that lack higher frequency details.
For cross-reenactment, our method achieves on average the best performance across most of the metrics.

\subsection{Ablation study}
We conduct multiple ablation studies to evaluate the efficacy of different components of our method.
\begin{figure}
    \centering
    \includegraphics[width=\linewidth]{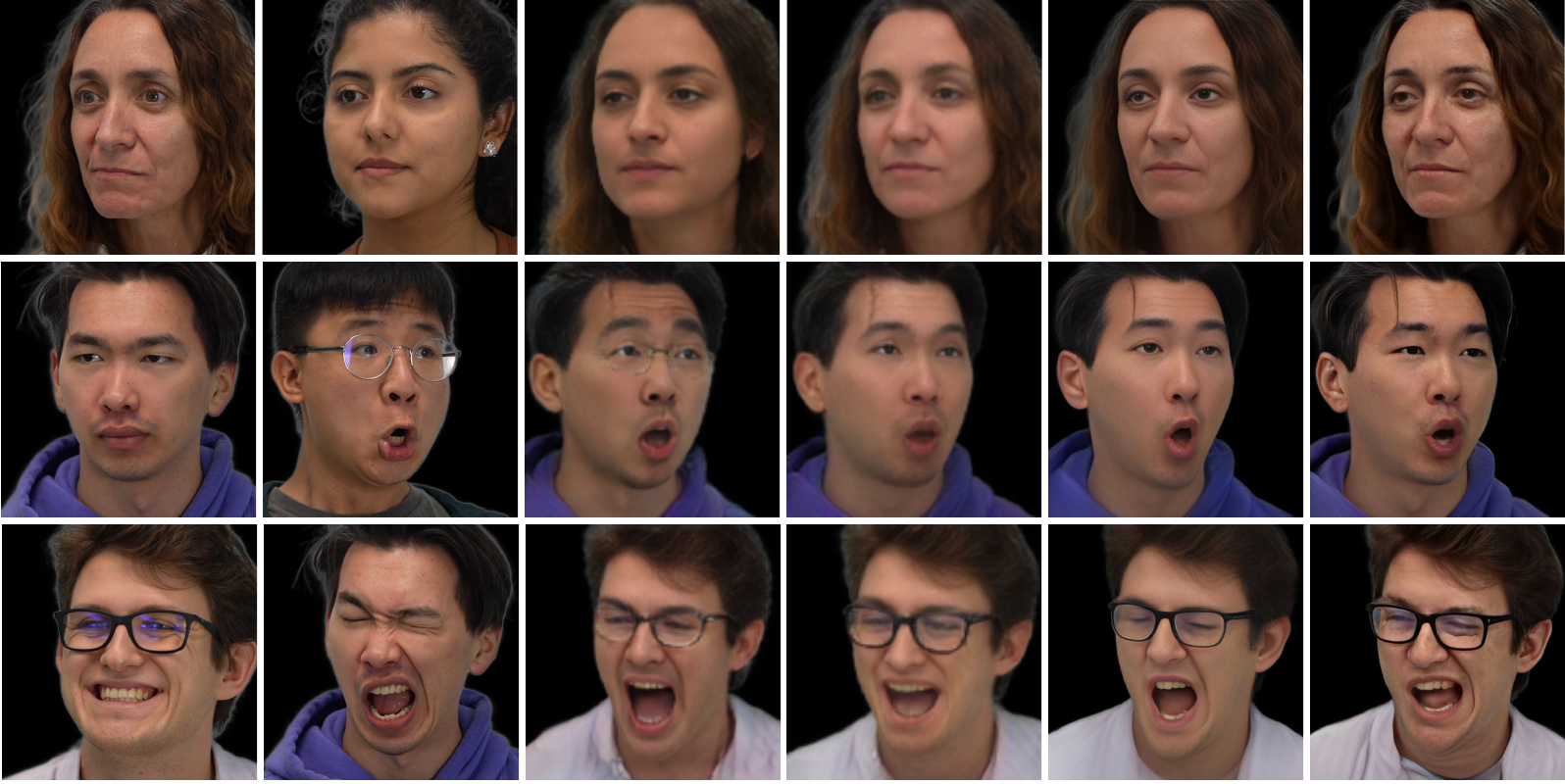}
    \newcolumntype{Y}{>{\centering\arraybackslash}X}
    \begin{tabularx}{\linewidth}{YYYYYY}
         \textbf{Source} & \textbf{Driver} & \footnotesize \textbf{w/o $\mathcal{L}_\text{neu}$} & \textbf{Stage 1} & \textbf{Stage 2} & \textbf{Stage 3}
    \end{tabularx}
    \vspace{-0.6cm}
    \caption{The quality, identity preservation, and expression are notably improved after each stage of our training pipeline}
    \label{fig:abl_stage123}
    \vspace{-0.1cm}
\end{figure}


\begin{figure}
    \centering
    \includegraphics[width=\linewidth]{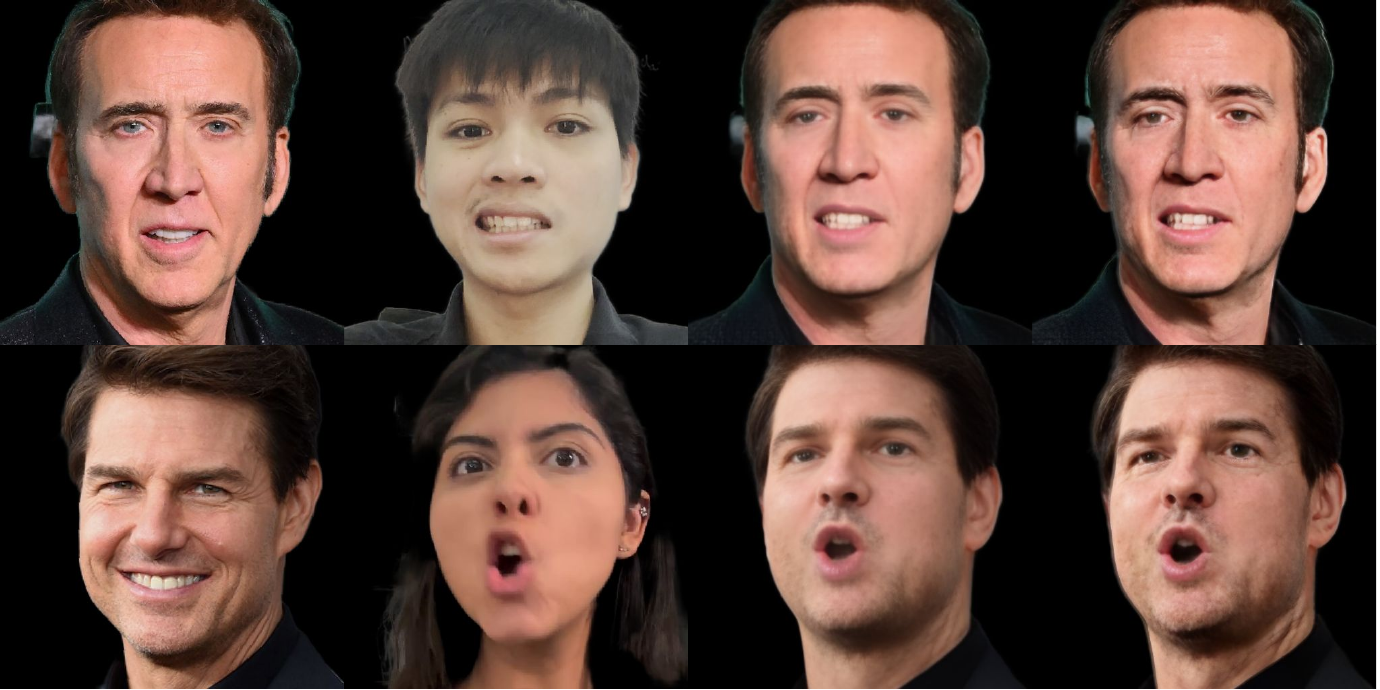}
    \newcolumntype{Y}{>{\centering\arraybackslash}X}
    \begin{tabularx}{\linewidth}{YYYY}
         \textbf{Source} & \textbf{Driver} & \textbf{w/o GAN} & \textbf{w GAN}
    \end{tabularx}
    \vspace{-0.6cm}
    \caption{GAN loss in Stage 3 adds high-frequency details to the results.}
    \label{fig:superres_eval}
    \vspace{-0.1cm}
\end{figure}

\begin{table}
    \centering
    \setlength{\tabcolsep}{3pt}
    \begin{tabular}{l|cccc|}
         Method & CSIM $\uparrow$ & FID $\downarrow$ & AED $\downarrow$ & APD $\downarrow$ \\
         \hline
         w/o $\mathcal{L}_{neu}$ & 0.29 & 98.49 & 4.03 & 0.059 \\
         Stage 1 & 0.60 & 88.88 & 5.68 & 0.08 \\
         Stage 2 & 0.58 & 38.96 & 5.25 & 0.07 \\
         Stage 3 & 0.67 & 25.80 & 6.31 & 0.09 \\
    \end{tabular}
    \caption{Quantitaitve ablation study on the use of neuralization loss and multi-stage training.}
    \vspace{-0.6cm}
    \label{tab:abl}
\end{table}
First, we note that without neutralizing loss, our method completely fails to disentangle between the identity and the expression, see quantitative results in Table~\ref{tab:abl} and qualitative in \cref{fig:abl_stage123}.
Moreover, each training stage consistently improves the overall performance of the method, with Stage 2 substantially improving the visual quality of the results (see FID metric) at the expense of reduced identity and expression preservation (CSIM, AED, and APD columns), while Stage 3 finally achieves the best performance across all metrics.
These findings are also validated by the qualitative results in \cref{fig:abl_stage123}.
\begin{figure}
    \centering
    \includegraphics[width=0.95\linewidth]{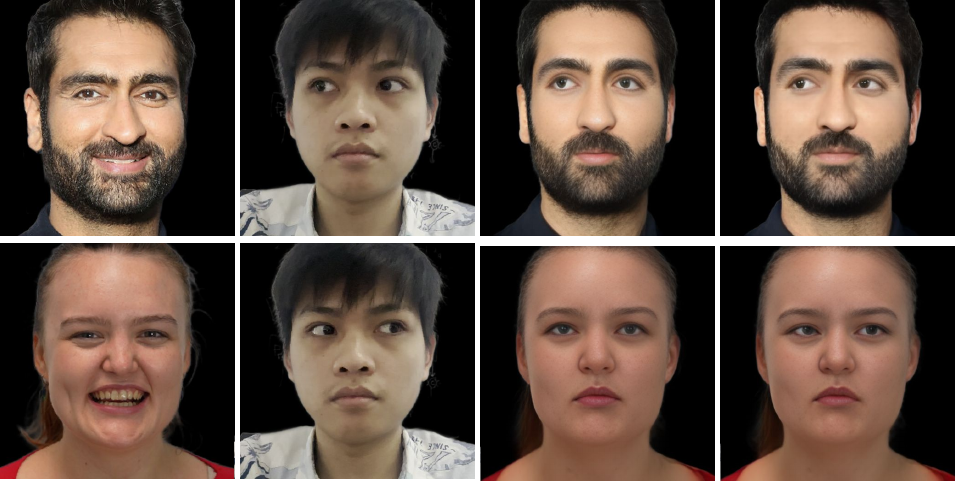}
    \newcolumntype{Y}{>{\centering\arraybackslash}X}
    \begin{tabularx}{0.95\linewidth}{YYYY}
         \textbf{Source} & \textbf{Driver} & \textbf{w/o eye loss} & \textbf{Ours}
    \end{tabularx}
    \vspace{-0.4cm}
    \caption{Our eye loss markedly improves the eye gaze accuracy. Digital zoom-in is recommended.}
    \label{fig:eye_qual}
    \vspace{-0.1cm}
\end{figure}
Then, we validate the effect of the separate eye gaze loss in \cref{fig:eye_qual} and find that without this loss the method fails to consistently reenact the correct eye gaze of the driver.
We also validate the effect of the GAN loss that we add in the third stage in \cref{fig:superres_eval} and show that it is highly effective in adding the high-frequency details into the reconstruction.
\begin{figure}
    \centering
    \includegraphics[width=0.95\linewidth]{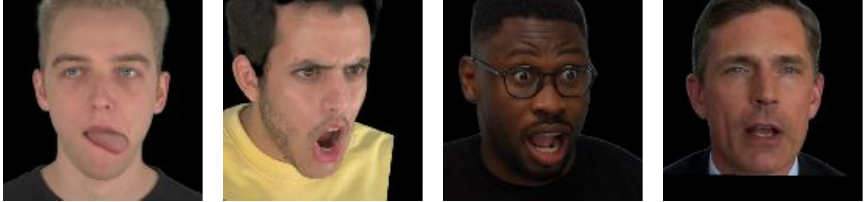} \\
    \vspace{-0.1cm}
    \textbf{Input} \\
    \vspace{0.05cm}
    \includegraphics[width=0.95\linewidth]{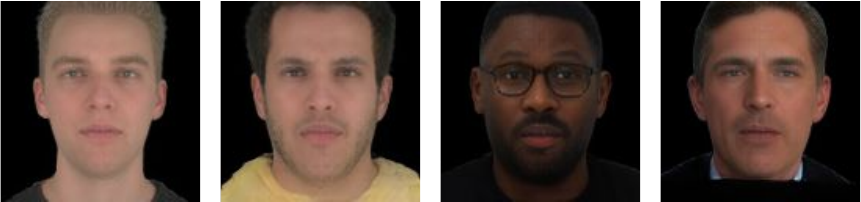} \\
    \vspace{-0.05cm}
    \textbf{Neutralized Image}
    \vspace{-0.3cm}
    \caption{Our neutralizer works for faces with tongue or extreme expressions.}
    \label{fig:neutralizer_qual}
    \vspace{-0.1cm}
\end{figure}
Lastly, we evaluate the effectiveness of our image neutralization network in \cref{fig:neutralizer_qual}.
We can see that it can consistently map the input into a shared neutral expression space, but introduces a minor identity change, which is the reason why we refrain from using it in Stages 2 and 3 of our training pipeline.

\subsection{Limitations}

\begin{figure}
    \centering
    \includegraphics[width=0.95\linewidth]{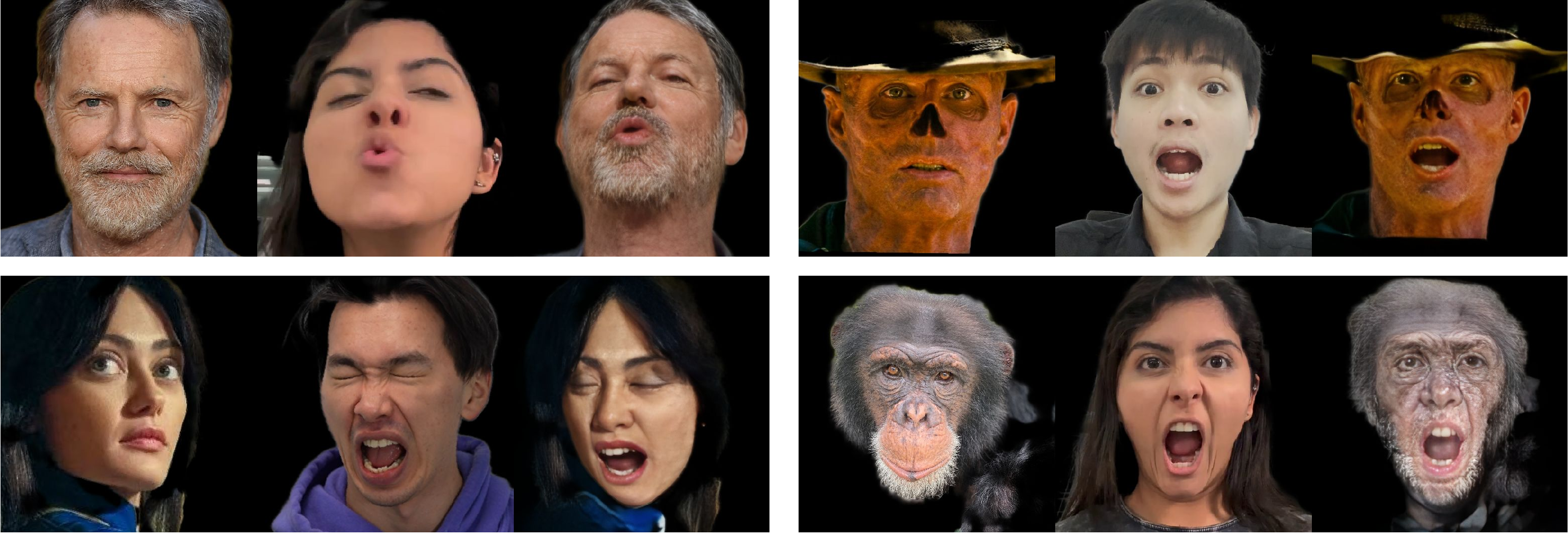}
    \vspace{-0.3cm}
    \caption{Our model's limitations include limited performance on out-of-distribution data, side views, and aliasing}
    \label{fig:limitations}
    \vspace{-0.1cm}
\end{figure}

Our method focuses on faithful reenactment of the facial region of the avatars.
One of the downsides of our approach is the lack of realistic hair modeling and rendering, which is manifested in aliasing artifacts in renders of fine hair structures, see \cref{fig:limitations}, top-left.
However, we note that these artifacts are also present in all the competing techniques.
While view-consistent, our method may also struggle to reenact the correct identity for extreme viewing angles and challenging lighting, see bottom-left example.
Lastly, our method hallucinates human features for sources with extreme makeups (the nose is still reenacted on the top-right example, even though it is not suppose to be in the source) or even ones that are animals, see bottom-right result.
%
%
%
%



\section{Conclusion}

We have presented a 3D aware one-shot neural head reenactment approach that can generate significantly more expressive facial expressions than the current state-of-the-art.
Our proposed VOODOO XP network suggests that such superior expressiveness is possible by directly transferring expression features of the driver to transformer blocks of the source's 3D lifting module. However, this approach only work if proper identity and expression disentanglement is possible for the expression transfer module, and the key is to develop an elaborate training procedure. We have demonstrated that an effective and fully volumetric disentanglement can be achieved using a multi-stage self-supervised learning approach, where explicit expression neutralization is combine with frontalized 3D lifting in the first stage, followed by a more relaxed but expressive fine-scale training step. A final global fine-tuning procedure using large video datasets, ensures a coverage of an extremely wide range of source identities.

During inference, our approach works without explicit expression neutralization for the 3D lifting of the source~\cite{deng2024portrait4dv2}, nor uses any explicit 3DMM for guiding the disentanglement~\cite{li2023generalizable}, and can handle unconstrained input portraits even with extreme expressions and head poses. Not only does our solution faithfully reproduce complex facial expressions from the driver and synthesize fine-scale dynamic skin deformations (e.g., wrinkle forming), but it can also handle extreme cases, such as squeezing faces, highly asymmetric expressions, and even tongues. Our experiments further show that our synthesized teeth and accessories (e.g., glasses) are more consistent and have less artifacts than several recent works. In particular, eyebrow movements behind glasses do not warp the frame of the glasses like many other methods. 

We have also introduced the first fully integrated VR telepresence solution for two-way immersive communication using an one-shot neural head reenactment technology. While recent works have shown how 3D aware head reenactment methods can be used for holographic displays~\cite{stengel2023,tran2023voodoo}, we demonstrate an immersive telepresence application using VR headsets. From a user experience stand point, our solution only requires a single input photo, which makes our solution significantly more accessible than current commercial solutions~\cite{applevisionpro} and research prototypes~\cite{DBLP:journals/corr/abs-1808-00362, Ma_2021_CVPR, 10.1145/3306346.3323030} for realistic avatar creation. Our system achieves full 30 fps at 512x512 resolution for stereoscopic VR rendering and can be already deployed, but portability is limited as it still relies on a VR HMD tethered to a powerful multi-GPU workstation for neural processing.

\paragraph{Ethics and Safety} Our reenactment method is intended to advance next generation immersive communication technologies, visual effects in movie production, and other entertainment applications. Like many other AI-based facial manipulation algorithms, our approach is highly accessible (single input image), and can be used to produce highly convincing synthetic content or allowing people to impersonate anyone in a VR environment. Hence, there is a risk for potential misuse of such technology, including the spread of disinformation, harassment, and fraud, which can lead to obvious harm and danger to society. We therefore prohibit and condemn all forms of malicious use of such technology, and are committed to raising awareness and educating the public about its danger. Furthermore, our models and data will be shared with the deepfake detection community to help combat malicious uses of AI generated videos.

\paragraph{Future Work} Our current one-shot head reenactment method only produces a head model and does not generate the back of it and the body of the person. For a fully immersive VR experience, we plan to incorporate recent techniques for 360 head generation~\cite{An_2023_CVPR} and body modeling into our framework. Furthermore, our real-time requirement is limiting the maximum resolution that can be generated by our neural radiance field representation on today’s hardware. Promising research on Gaussian splat-based avatars~\cite{qian2023gaussianavatars,xu2023gaussianheadavatar,saito2023relightable} are possible directions for future exploration.
\bibliographystyle{ACM-Reference-Format}
\bibliography{samples/sample-base}
In this supplementary material, we provide the additional results including more evaluations and comparisons.

\section{Appendix}
\subsection{Multi-view and Light Variant Evaluation}
\begin{figure}
    \centering
    \includegraphics[width=0.95\linewidth]{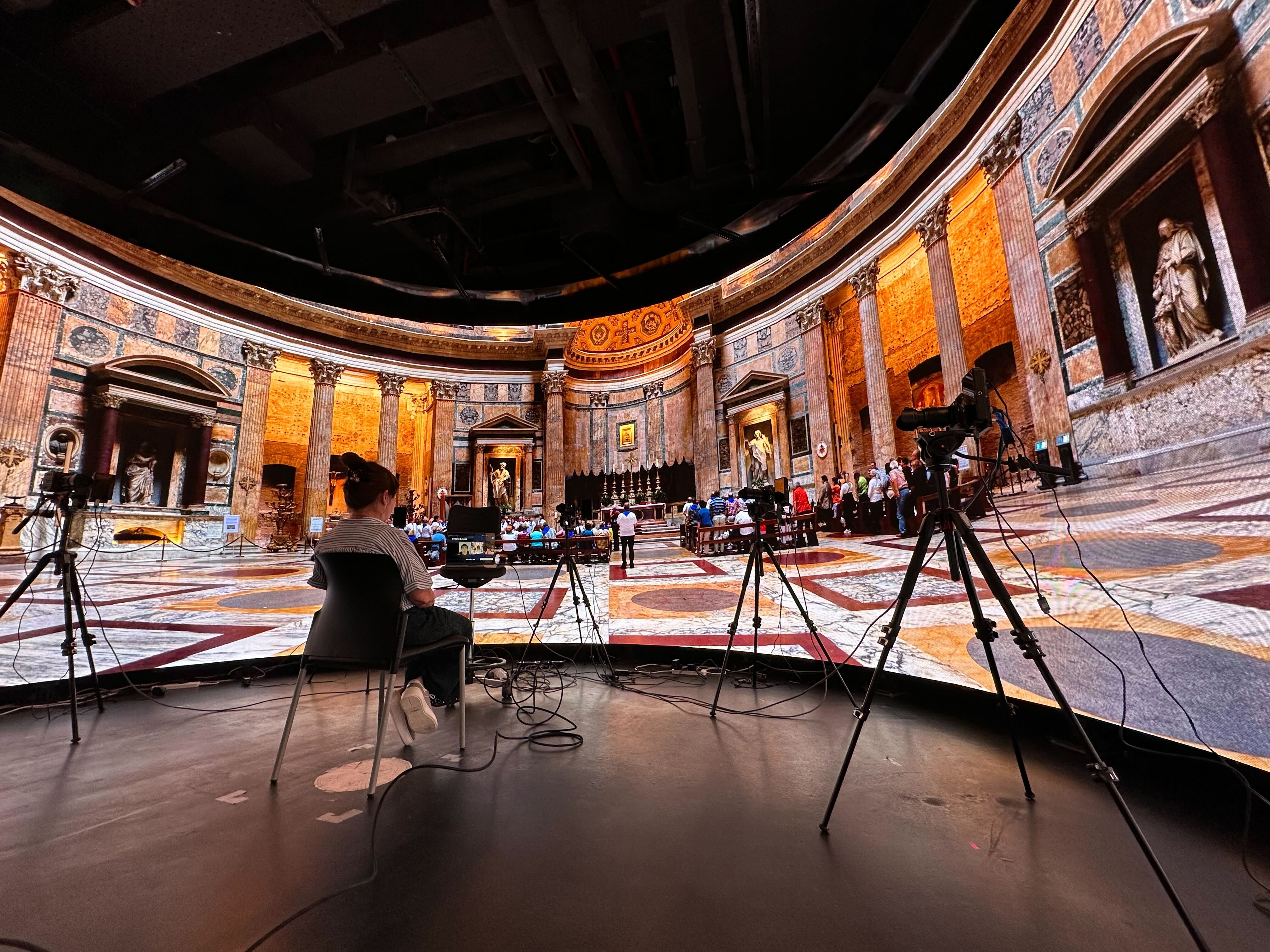}
    \caption{Our multi-view capture setup with controlled lighting.}
    \label{fig:metawall}
\end{figure}

\begin{figure*}
     \centering
     \includegraphics[width=0.95\linewidth]{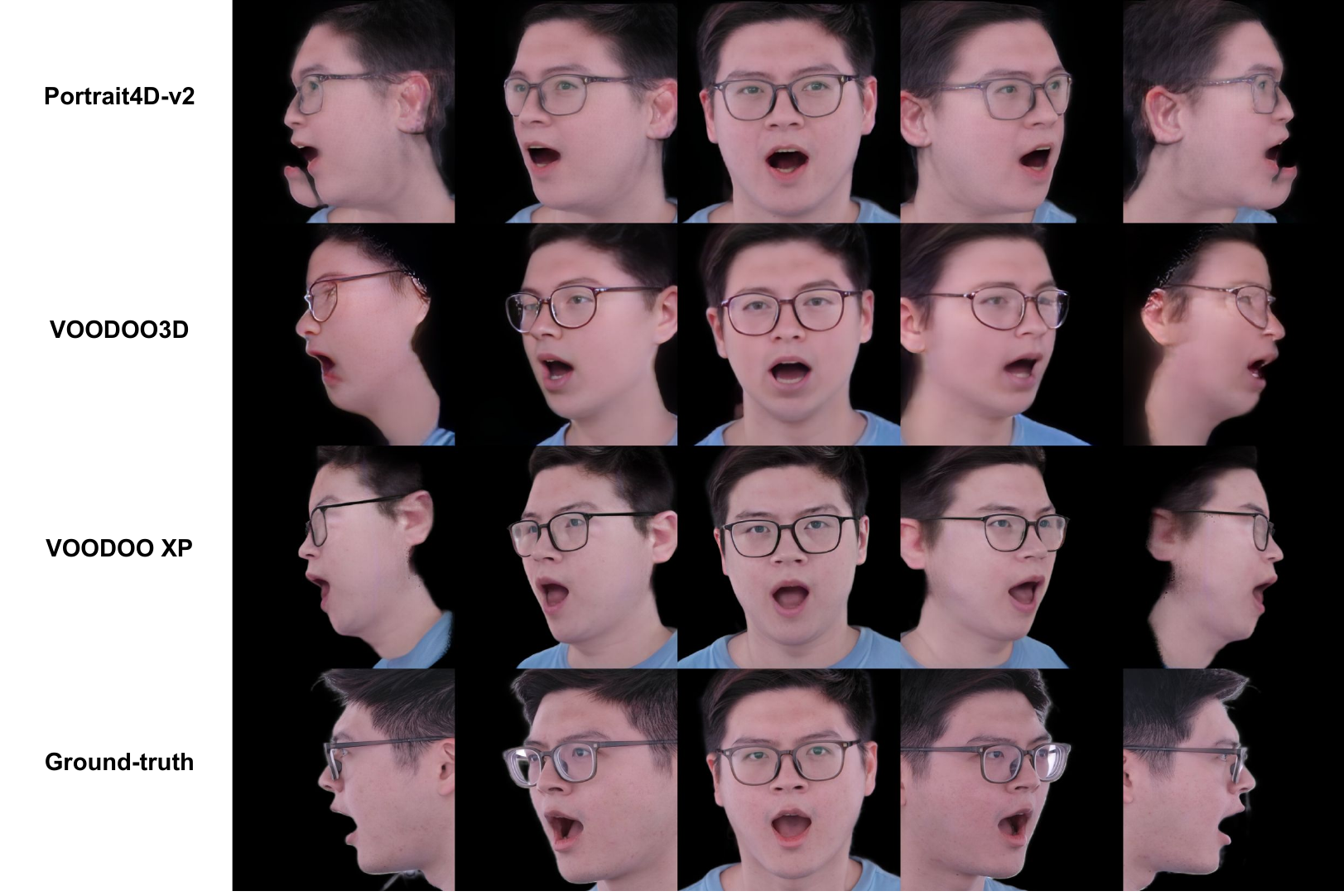}
     \caption{Self-reenactment results on our multi-view dataset with diverse light variant. For video result, refer to the attached video}
     \label{fig:meta01}
 \end{figure*}

\begin{figure*}
     \centering
     \includegraphics[width=0.95\linewidth]{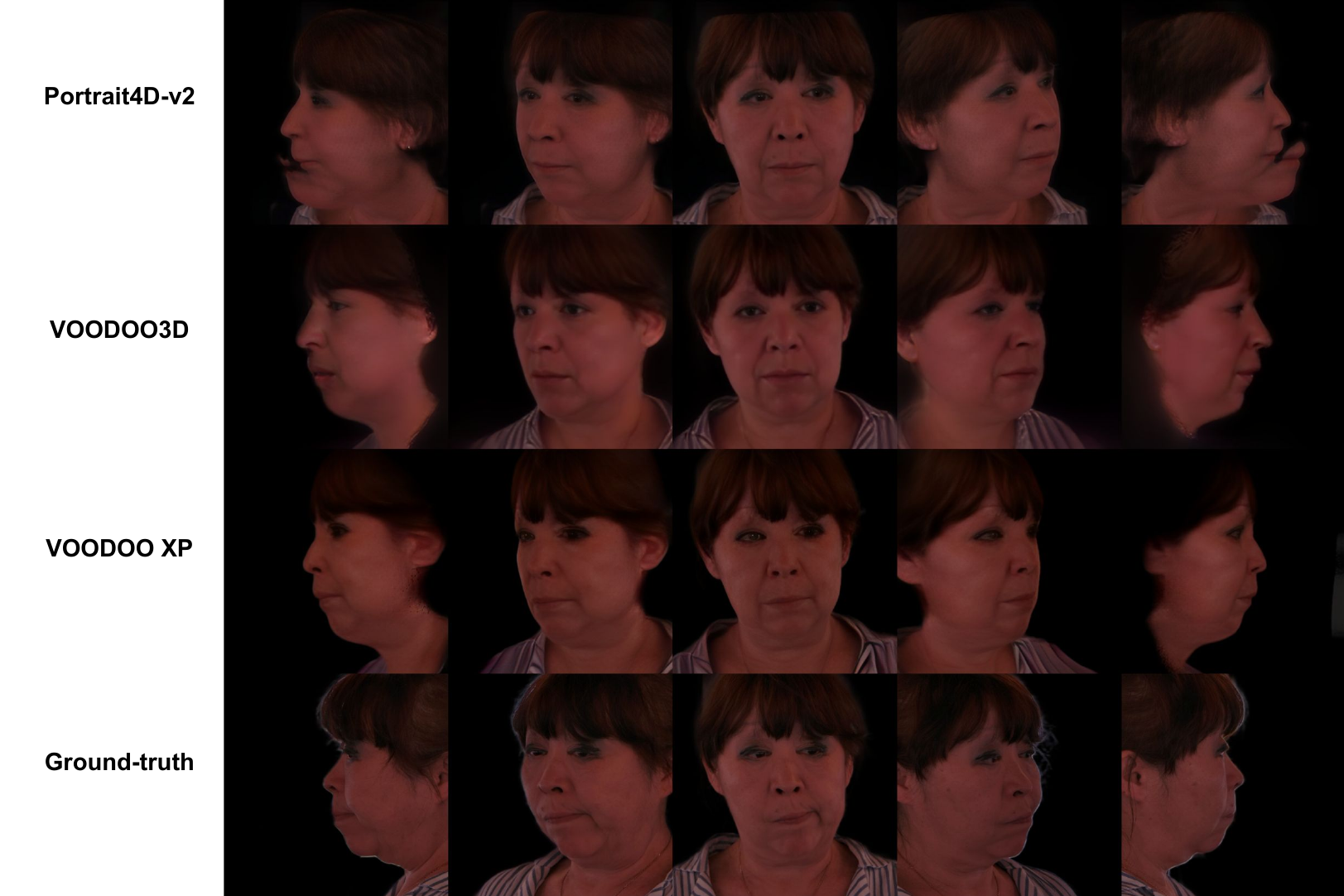}
     \caption{Self-reenactment results on our multi-view dataset with diverse light variant. For video result, refer to the attached video}
     \label{fig:meta02}
 \end{figure*}

\begin{figure*}
     \centering
     \includegraphics[width=0.95\linewidth]{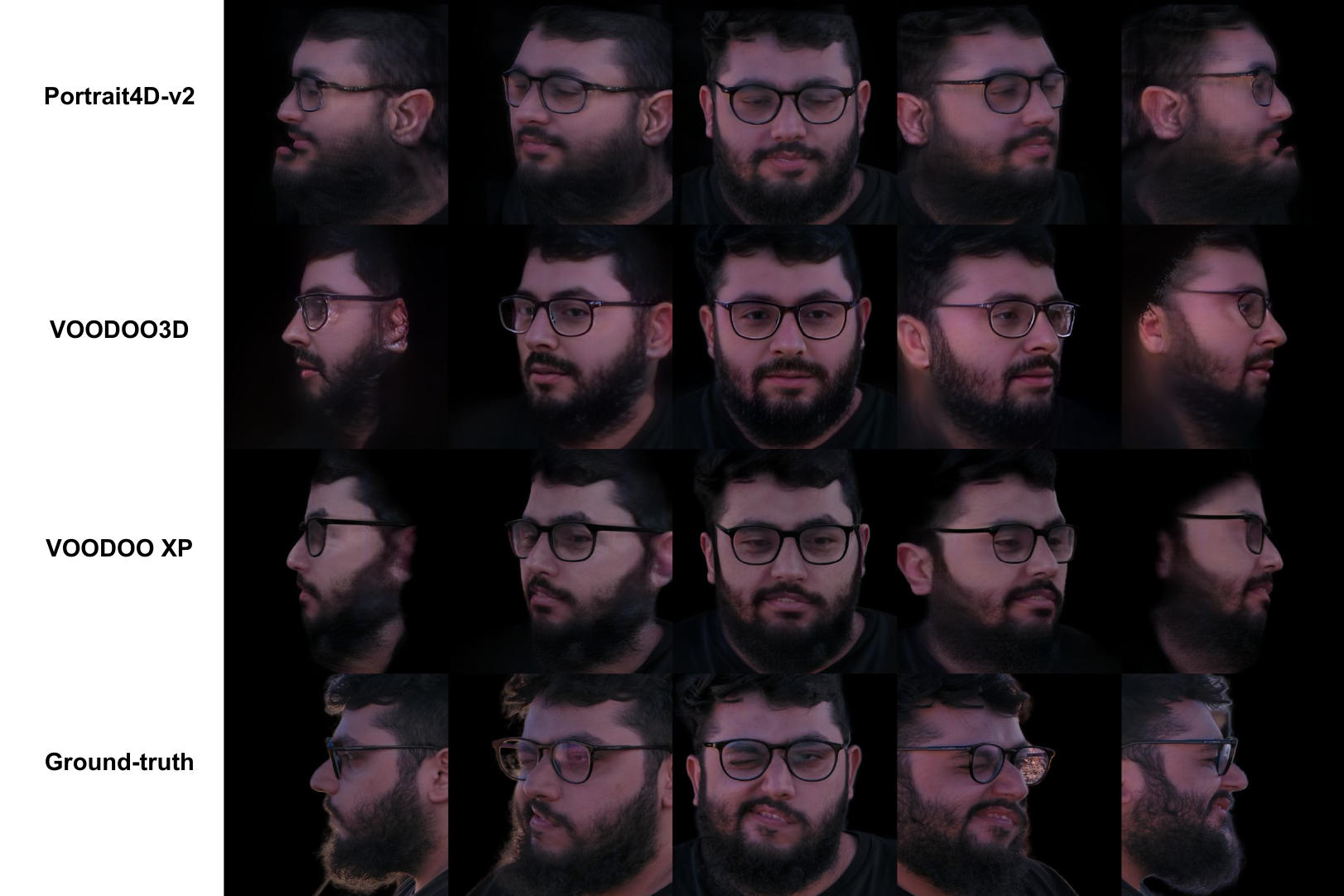}
     \caption{Self-reenactment results on our multi-view dataset with diverse light variant. For video result, refer to the attached video}
     \label{fig:meta03}
 \end{figure*}

To evaluate our method's multi-view consistency under various lighting conditions, we build a multi-view dataset using the system shown in \cref{fig:metawall}. The system comprises five Z CAM E2 cameras, each capturing videos at 1080p and 60FPS. These cameras are synchronized by a Blackmagic Design ATEM Mini Extreme ISO. The whole system is placed inside a 270-degree curved LED screen with a diameter of 5.7 meters, which can simulate different lighting conditions. Each sequence in the collected data includes five fixed viewpoints of the subject, providing ground truth to measure the novel view synthesis capability of 3D-aware facial reenactment methods. In \cref{fig:meta01}, \cref{fig:meta02}, and \cref{fig:meta03}, we compare Portrait4D-v2 \cite{deng2024portrait4dv2}, VOODOO 3D \cite{tran2023voodoo}, and our method on the self-reenactment task using this dataset. Our model shows better identity preservation than other methods and produces consistent novel views even under extreme lighting conditions. For video result of these examples, please refer to our attached video.

 \subsection{Additional Self-reenactment Qualitative Results}
\begin{figure*}[ht]
     \centering
     \includegraphics[width=0.7\linewidth]{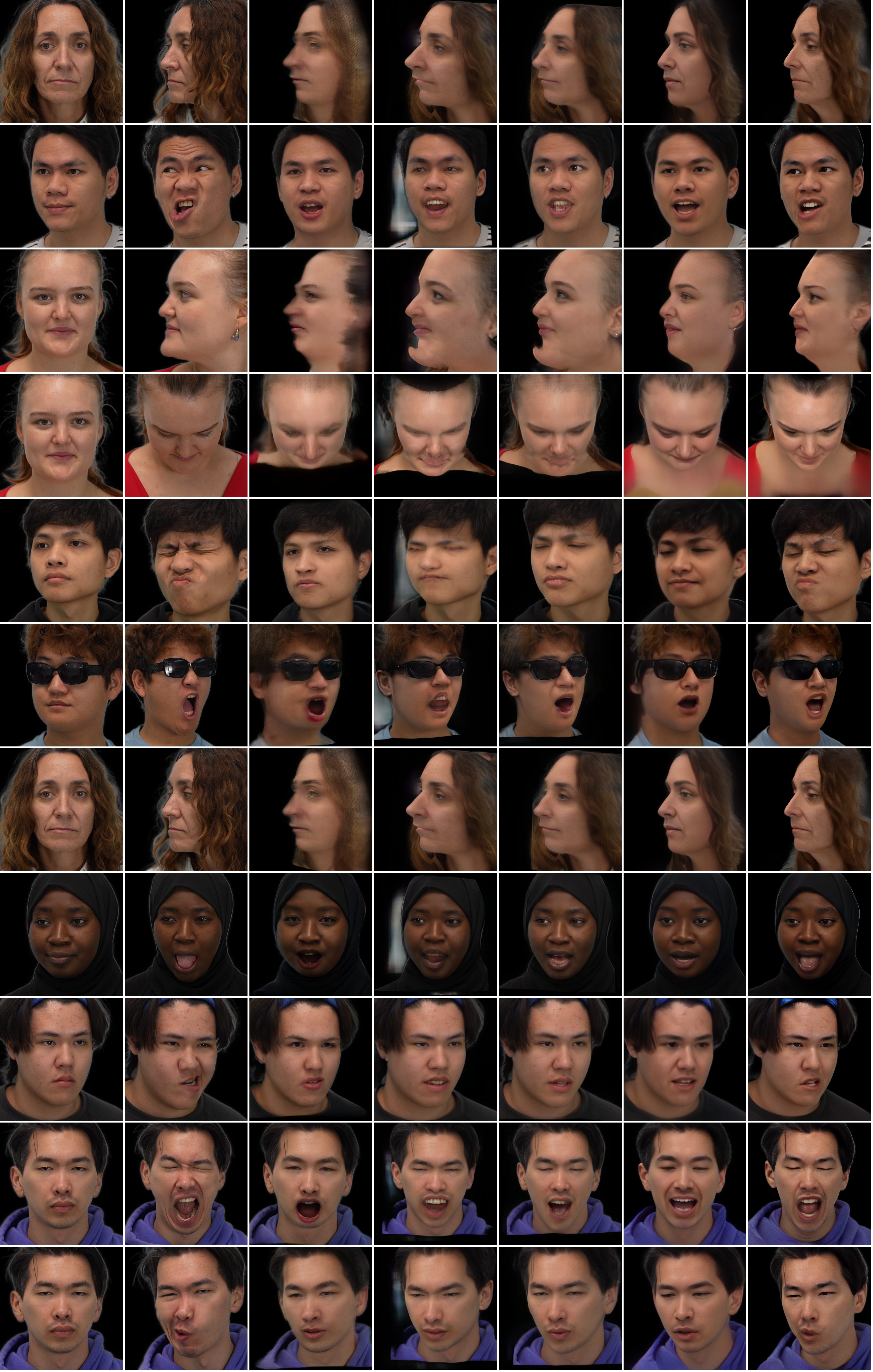}
    \newcolumntype{Y}{>{\centering\arraybackslash}X}
    \begin{tabularx}{0.7\linewidth}{YYYYYYY}
         Source & Driver & GOHA & Portrait4D & P4D-v2 & VOODOO3D & Ours
    \end{tabularx}
     \caption{Additional self-reenactment comparisons}
     \label{fig:selfqual}
 \end{figure*}

 Additional comparisons between our model with other state-of-the-art methods on self-reenactment task as provided. Results are given in \cref{fig:selfqual}.

 \subsection{Additional Cross-reenactment Qualitative Comparisons}
\begin{figure*}[ht]
     \centering
     \includegraphics[width=0.95\linewidth]{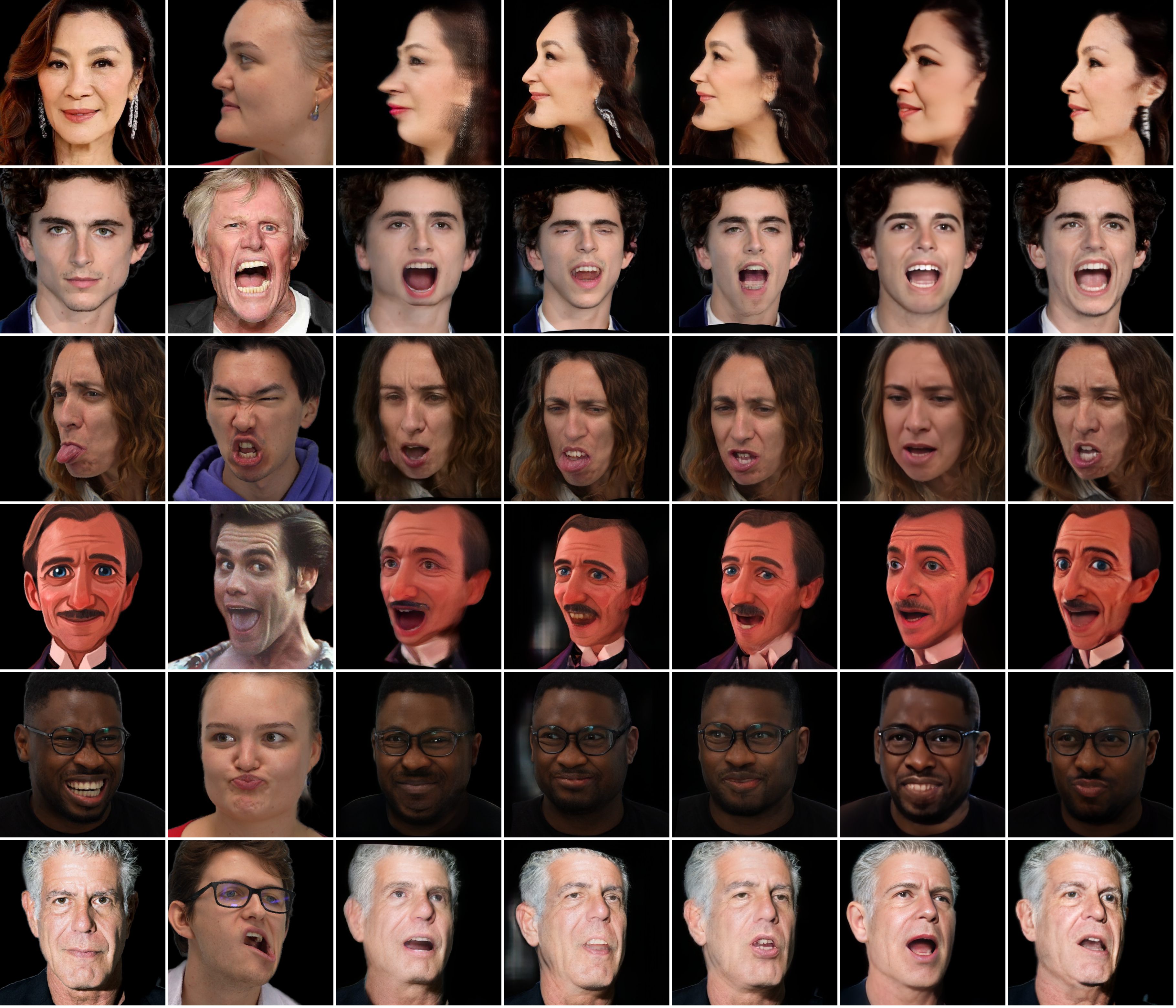}
    \newcolumntype{Y}{>{\centering\arraybackslash}X}
    \begin{tabularx}{0.95\linewidth}{YYYYYYY}
         Source & Driver & GOHA & Portrait4D & P4D-v2 & VOODOO3D & Ours
    \end{tabularx}
     \caption{Additional qualitative comparisons}
     \label{fig:addcom}
 \end{figure*}

Additional qualitative cross-reenactment comparisons are shown in \cref{fig:addcom}.
 
 \subsection{Additional Cross-reenactment Results}
\begin{figure*}
     \centering
     \includegraphics[width=0.95\linewidth]{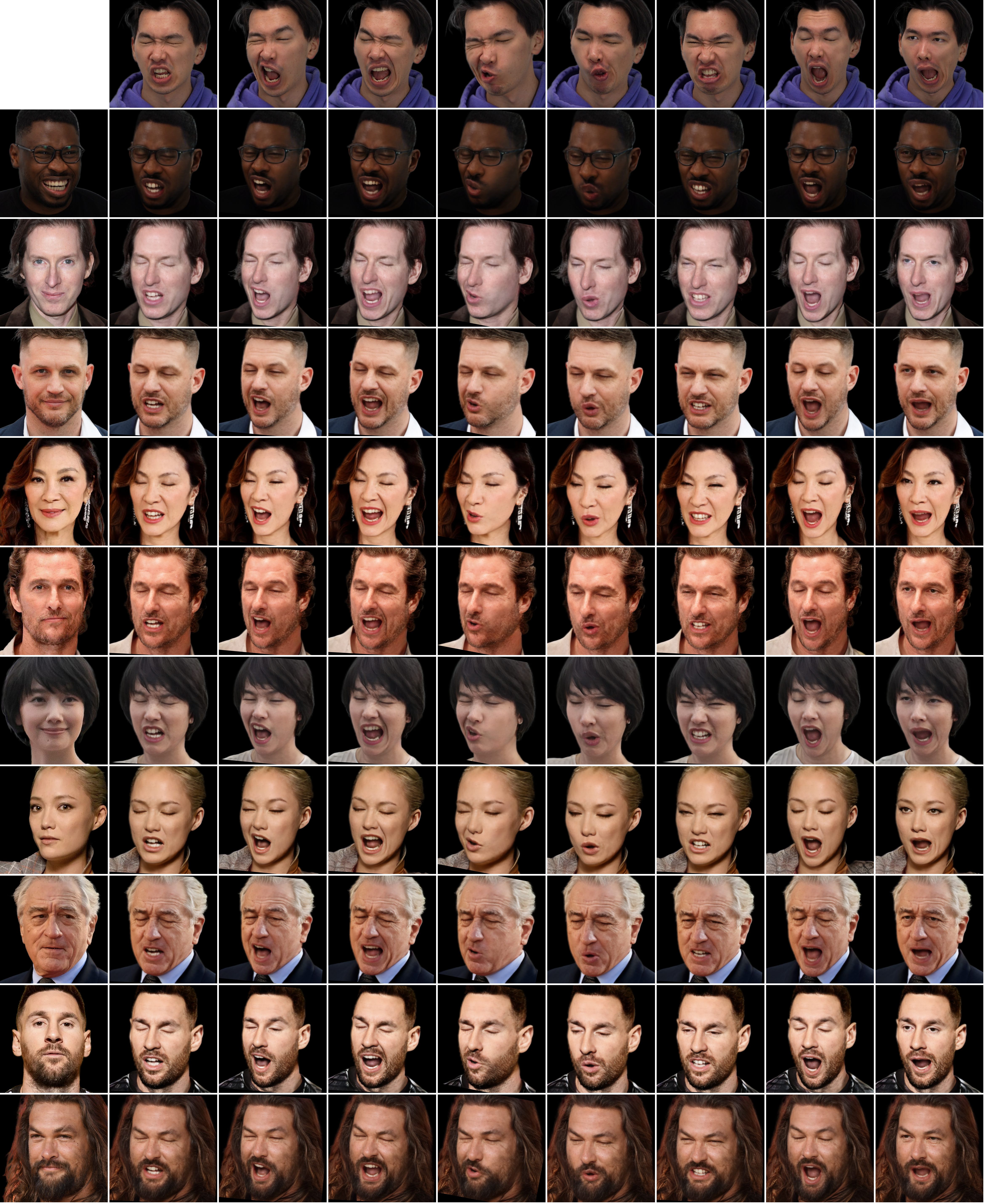}
     \caption{Additional cross-reenactment results of our method}
     \label{fig:addqual01}
 \end{figure*}

\begin{figure*}
     \centering
     \includegraphics[width=0.95\linewidth]{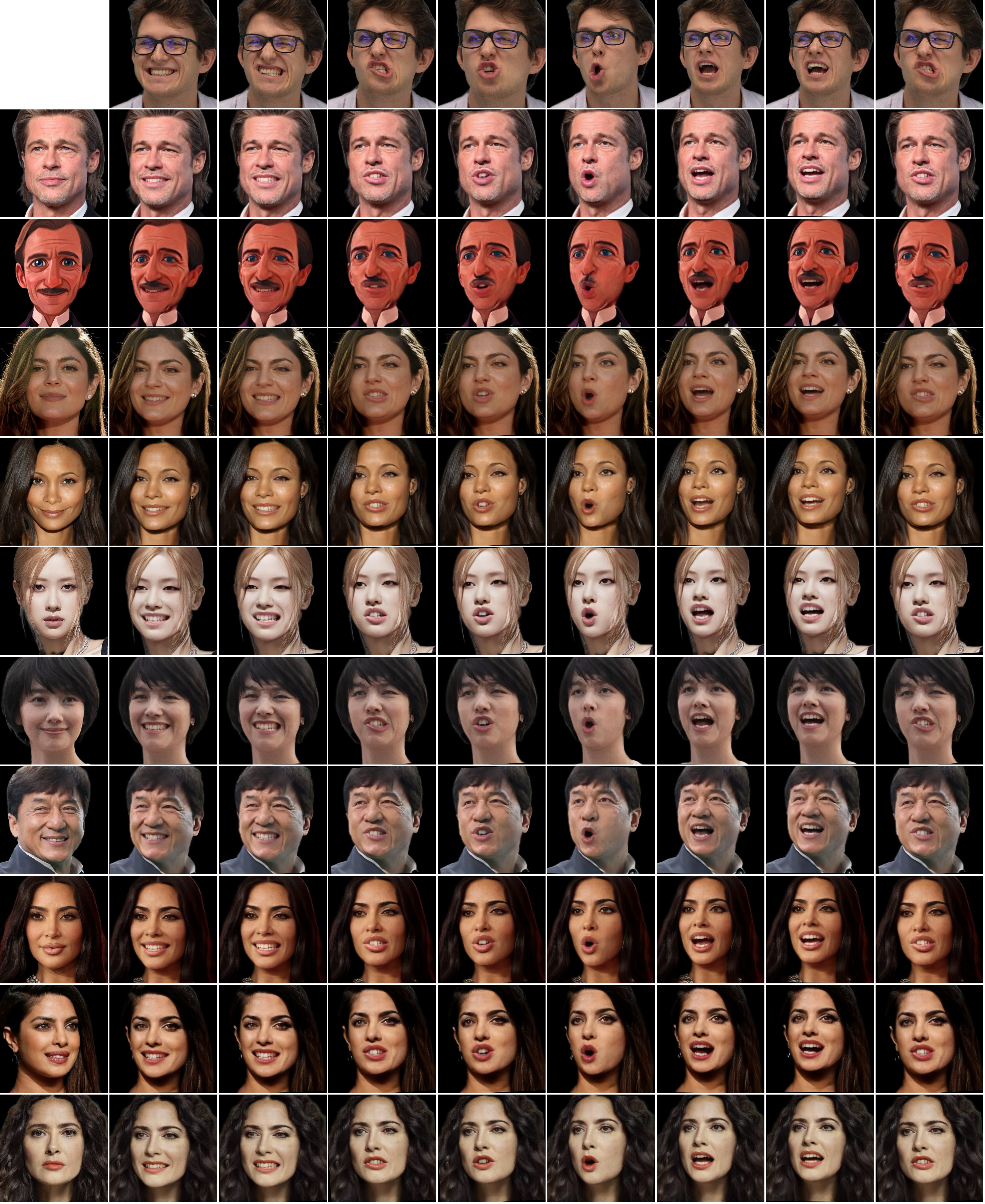}
     \caption{Additional cross-reenactment results of our method}
     \label{fig:addqual02}
 \end{figure*}

\begin{figure*}
     \centering
     \includegraphics[width=0.95\linewidth]{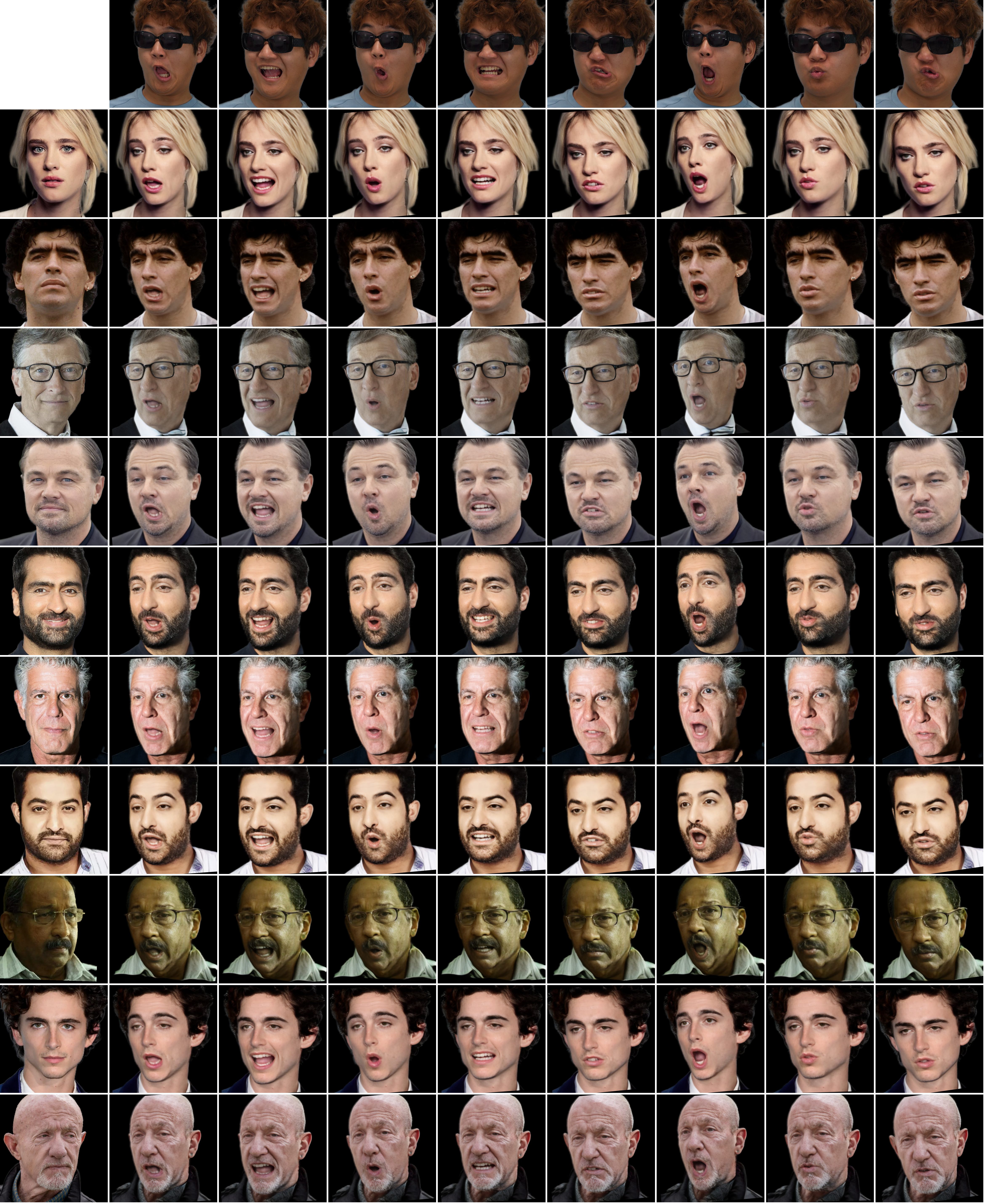}
     \caption{Additional cross-reenactment results of our method}
     \label{fig:addqual03}
 \end{figure*}

We provide 240 additional cross-reenactment results of our method in \cref{fig:addqual01}, \cref{fig:addqual02}, and \cref{fig:addqual03}. For each figure, the first column consists of the sources and the first row are the drivers. Each cell $(i, j)$ is the cross-reenactment result of the $i^{th}$ source and $j^{th}$ driver.

\end{document}


\title{Supplementary Material \protect\\ VOODOO XP: Expressive One-Shot Head Reenactment for VR Telepresence}

\author{Ben Trovato}
\authornote{Both authors contributed equally to this research.}
\email{trovato@corporation.com}
\orcid{1234-5678-9012}
\author{G.K.M. Tobin}
\authornotemark[1]
\email{webmaster@marysville-ohio.com}
\affiliation{%
  \institution{Institute for Clarity in Documentation}
  \city{Dublin}
  \state{Ohio}
  \country{USA}
}

\author{Lars Th{\o}rv{\"a}ld}
\affiliation{%
  \institution{The Th{\o}rv{\"a}ld Group}
  \city{Hekla}
  \country{Iceland}}
\email{larst@affiliation.org}

\author{Valerie B\'eranger}
\affiliation{%
  \institution{Inria Paris-Rocquencourt}
  \city{Rocquencourt}
  \country{France}
}

\author{Aparna Patel}
\affiliation{%
 \institution{Rajiv Gandhi University}
 \city{Doimukh}
 \state{Arunachal Pradesh}
 \country{India}}

\author{Huifen Chan}
\affiliation{%
  \institution{Tsinghua University}
  \city{Haidian Qu}
  \state{Beijing Shi}
  \country{China}}

\author{Charles Palmer}
\affiliation{%
  \institution{Palmer Research Laboratories}
  \city{San Antonio}
  \state{Texas}
  \country{USA}}
\email{cpalmer@prl.com}

\author{John Smith}
\affiliation{%
  \institution{The Th{\o}rv{\"a}ld Group}
  \city{Hekla}
  \country{Iceland}}
\email{jsmith@affiliation.org}

\author{Julius P. Kumquat}
\affiliation{%
  \institution{The Kumquat Consortium}
  \city{New York}
  \country{USA}}
\email{jpkumquat@consortium.net}

\renewcommand{\shortauthors}{Trovato et al.}

\ccsdesc[500]{Computing methodologies~Image-based rendering}

\keywords{View Synthesis, Facial Animation, Neural Radiance Field}



\maketitle
In this supplementary material, we provide the additional results including more evaluations and comparisons.

\section{Appendix}
\subsection{Multi-view and Light Variant Evaluation}
\begin{figure}
    \centering
    \includegraphics[width=0.95\linewidth]{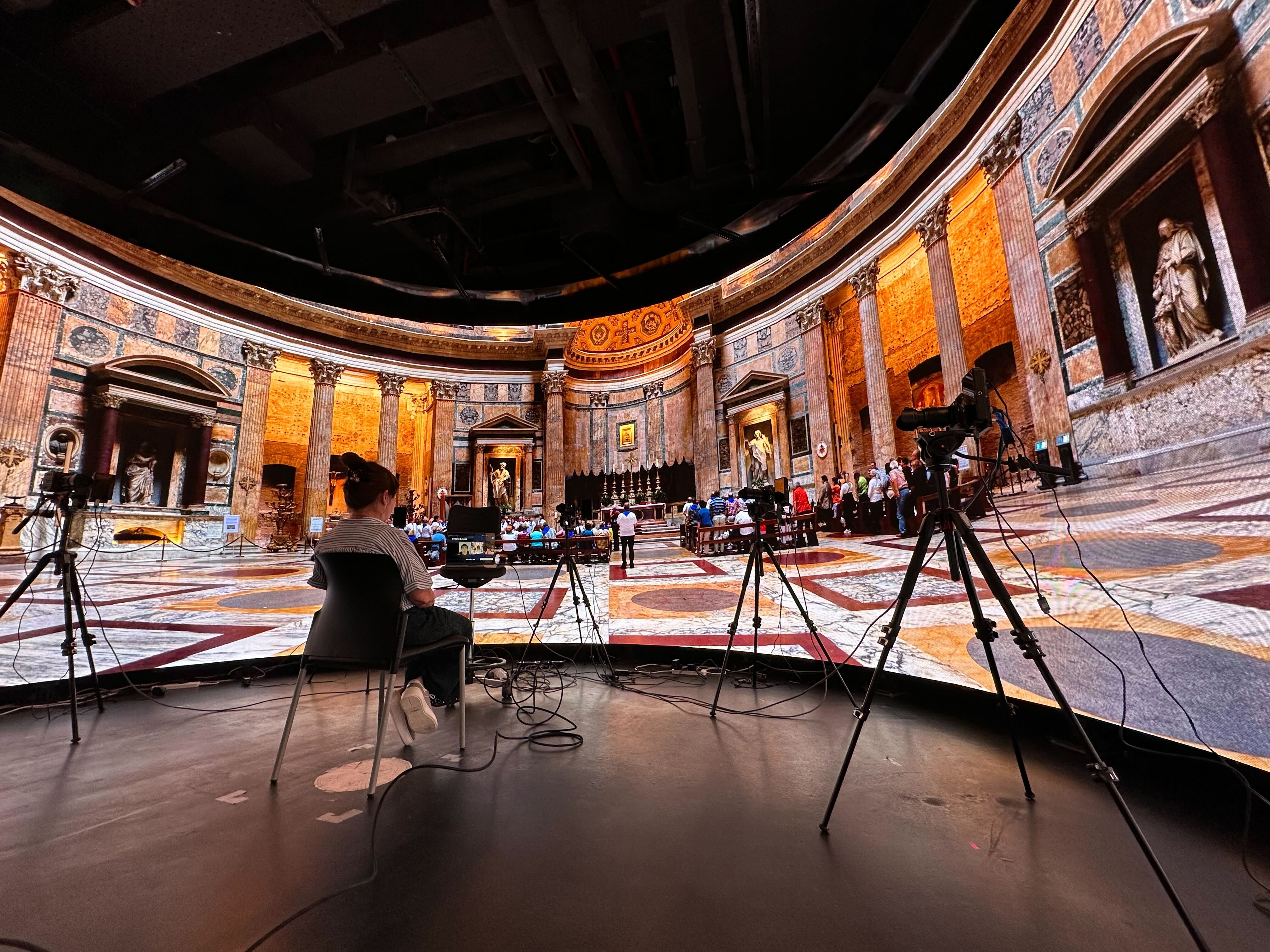}
    \caption{Our multi-view capture setup with controlled lighting.}
    \label{fig:metawall}
\end{figure}

\begin{figure*}
     \centering
     \includegraphics[width=0.95\linewidth]{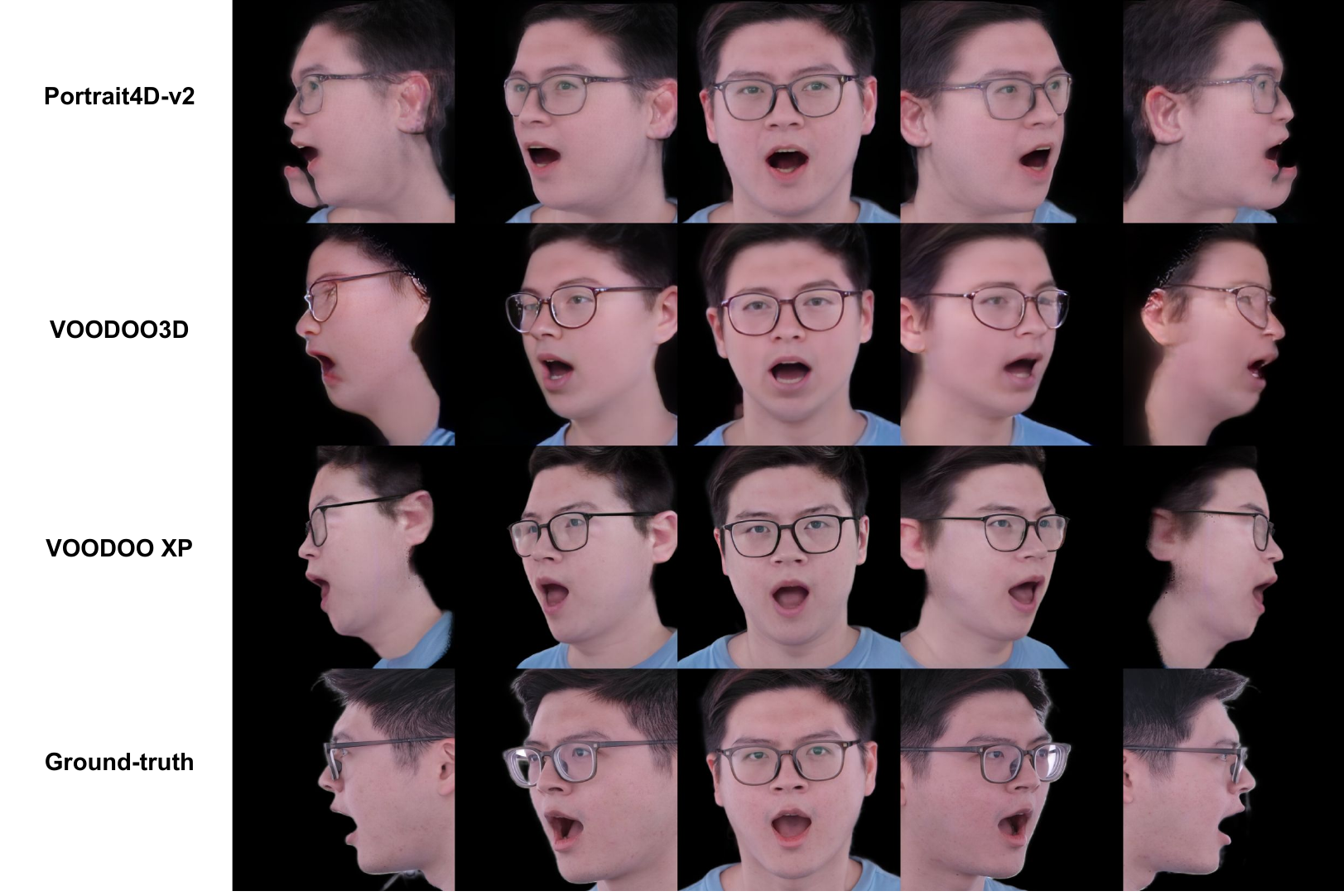}
     \caption{Self-reenactment results on our multi-view dataset with diverse light variant. For video result, refer to the attached video}
     \label{fig:meta01}
 \end{figure*}

\begin{figure*}
     \centering
     \includegraphics[width=0.95\linewidth]{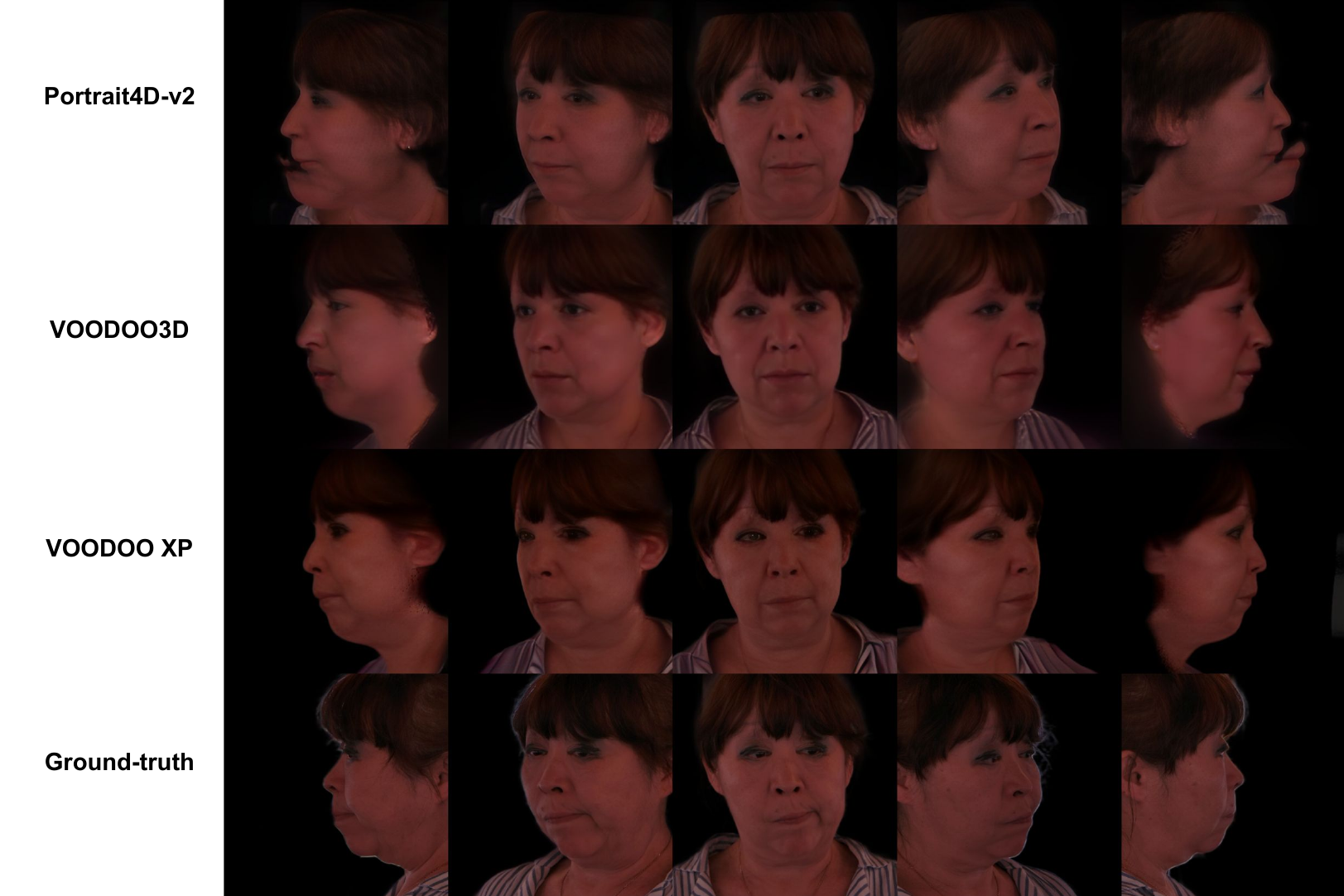}
     \caption{Self-reenactment results on our multi-view dataset with diverse light variant. For video result, refer to the attached video}
     \label{fig:meta02}
 \end{figure*}

\begin{figure*}
     \centering
     \includegraphics[width=0.95\linewidth]{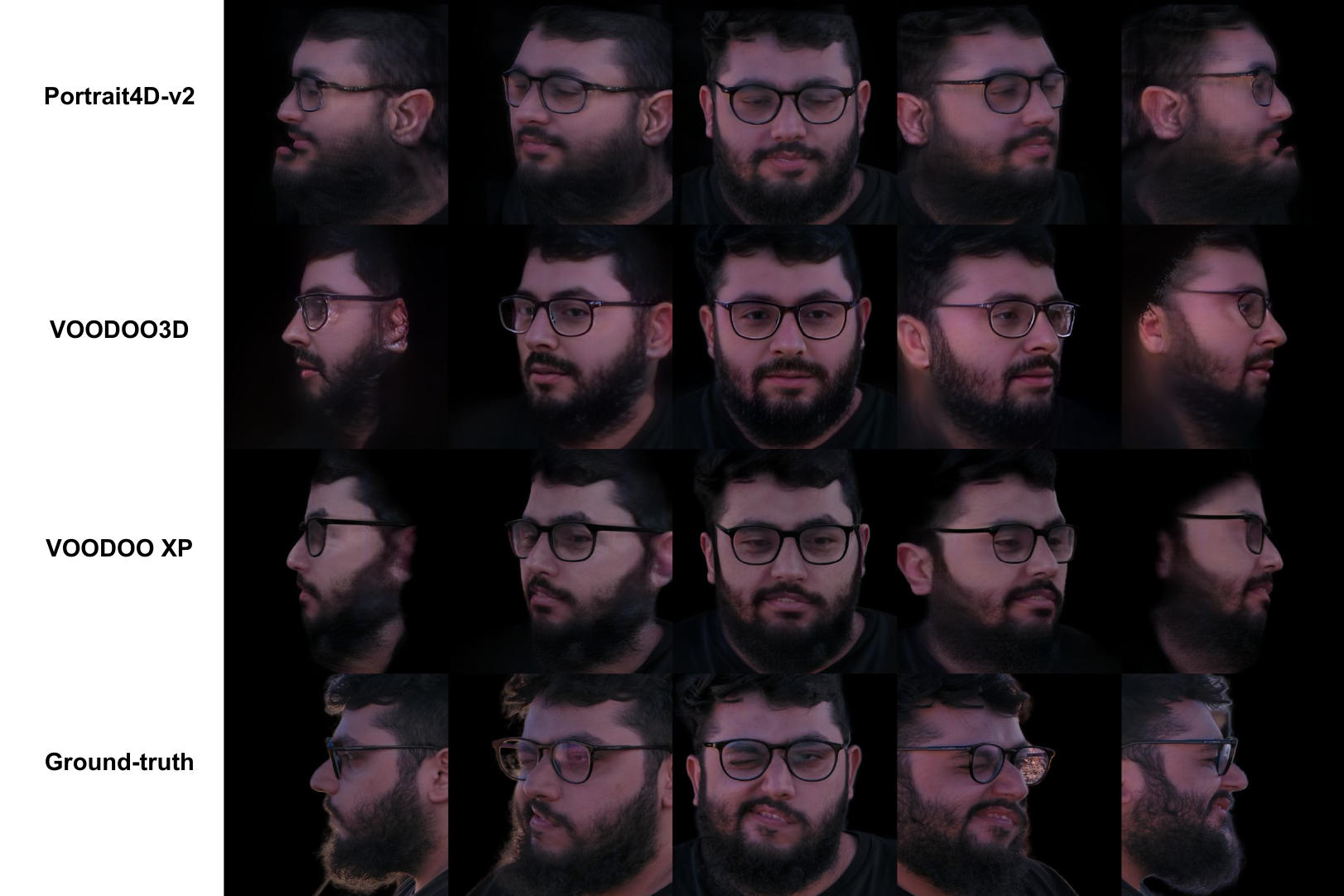}
     \caption{Self-reenactment results on our multi-view dataset with diverse light variant. For video result, refer to the attached video}
     \label{fig:meta03}
 \end{figure*}

To evaluate our method's multi-view consistency under various lighting conditions, we build a multi-view dataset using the system shown in \cref{fig:metawall}. The system comprises five Z CAM E2 cameras, each capturing videos at 1080p and 60FPS. These cameras are synchronized by a Blackmagic Design ATEM Mini Extreme ISO. The whole system is placed inside a 270-degree curved LED screen with a diameter of 5.7 meters, which can simulate different lighting conditions. Each sequence in the collected data includes five fixed viewpoints of the subject, providing ground truth to measure the novel view synthesis capability of 3D-aware facial reenactment methods. In \cref{fig:meta01}, \cref{fig:meta02}, and \cref{fig:meta03}, we compare Portrait4D-v2 \cite{deng2024portrait4dv2}, VOODOO 3D \cite{tran2023voodoo}, and our method on the self-reenactment task using this dataset. Our model shows better identity preservation than other methods and produces consistent novel views even under extreme lighting conditions. For video result of these examples, please refer to our attached video.

 \subsection{Additional Self-reenactment Qualitative Results}
\begin{figure*}[ht]
     \centering
     \includegraphics[width=0.7\linewidth]{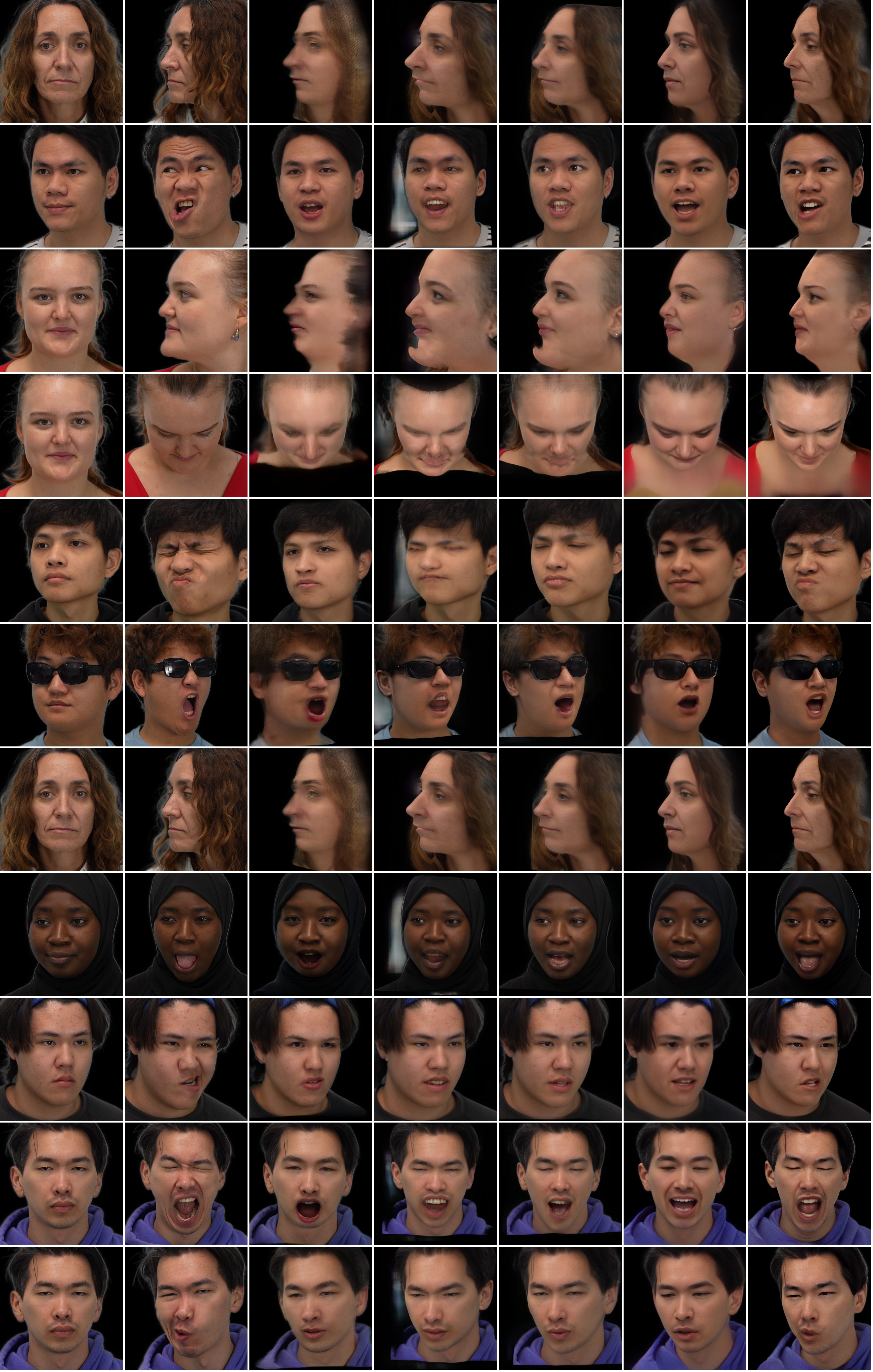}
    \newcolumntype{Y}{>{\centering\arraybackslash}X}
    \begin{tabularx}{0.7\linewidth}{YYYYYYY}
         Source & Driver & GOHA & Portrait4D & P4D-v2 & VOODOO3D & Ours
    \end{tabularx}
     \caption{Additional self-reenactment comparisons}
     \label{fig:selfqual}
 \end{figure*}

 Additional comparisons between our model with other state-of-the-art methods on self-reenactment task as provided. Results are given in \cref{fig:selfqual}.

 \subsection{Additional Cross-reenactment Qualitative Comparisons}
\begin{figure*}[ht]
     \centering
     \includegraphics[width=0.95\linewidth]{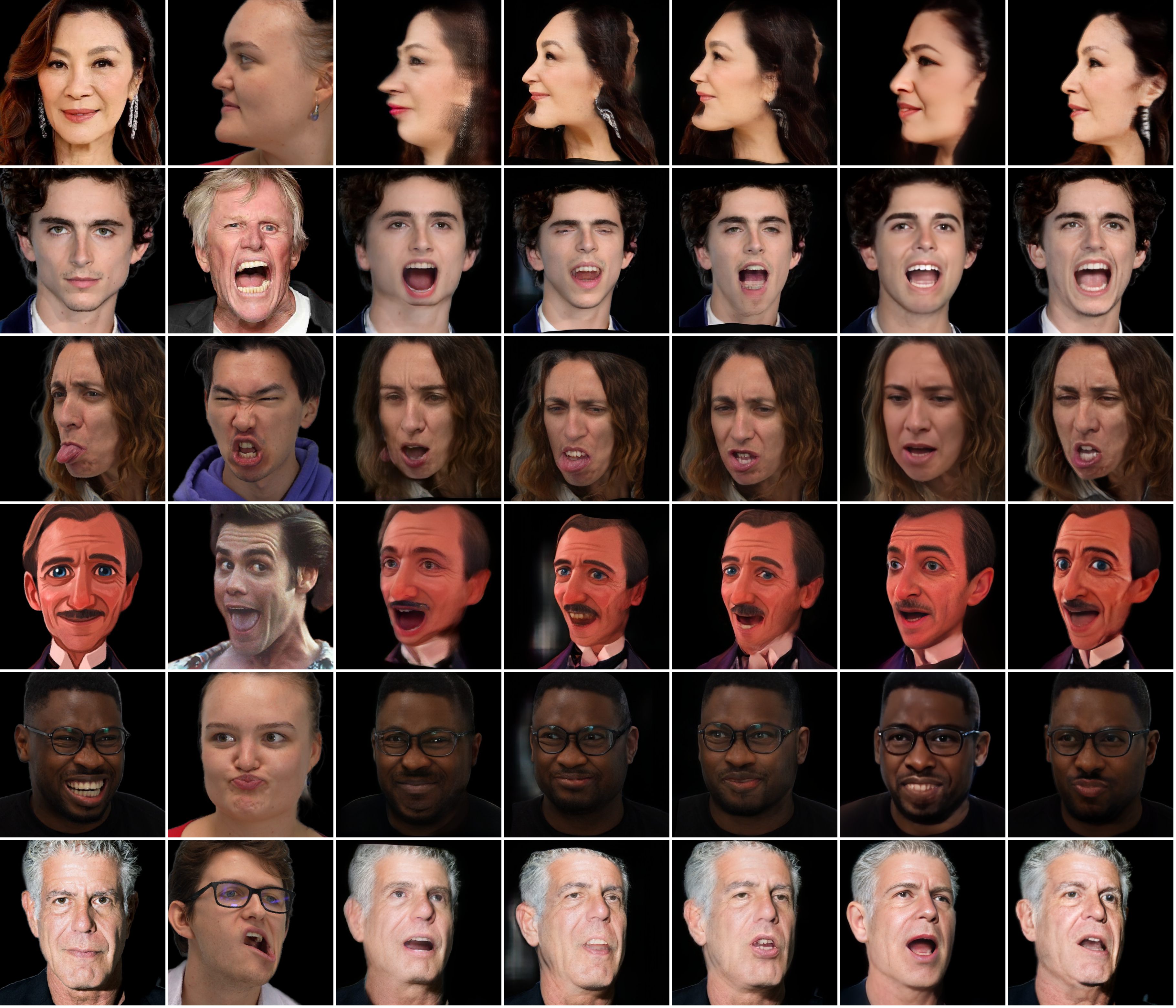}
    \newcolumntype{Y}{>{\centering\arraybackslash}X}
    \begin{tabularx}{0.95\linewidth}{YYYYYYY}
         Source & Driver & GOHA & Portrait4D & P4D-v2 & VOODOO3D & Ours
    \end{tabularx}
     \caption{Additional qualitative comparisons}
     \label{fig:addcom}
 \end{figure*}

Additional qualitative cross-reenactment comparisons are shown in \cref{fig:addcom}.
 
 \subsection{Additional Cross-reenactment Results}
\begin{figure*}
     \centering
     \includegraphics[width=0.95\linewidth]{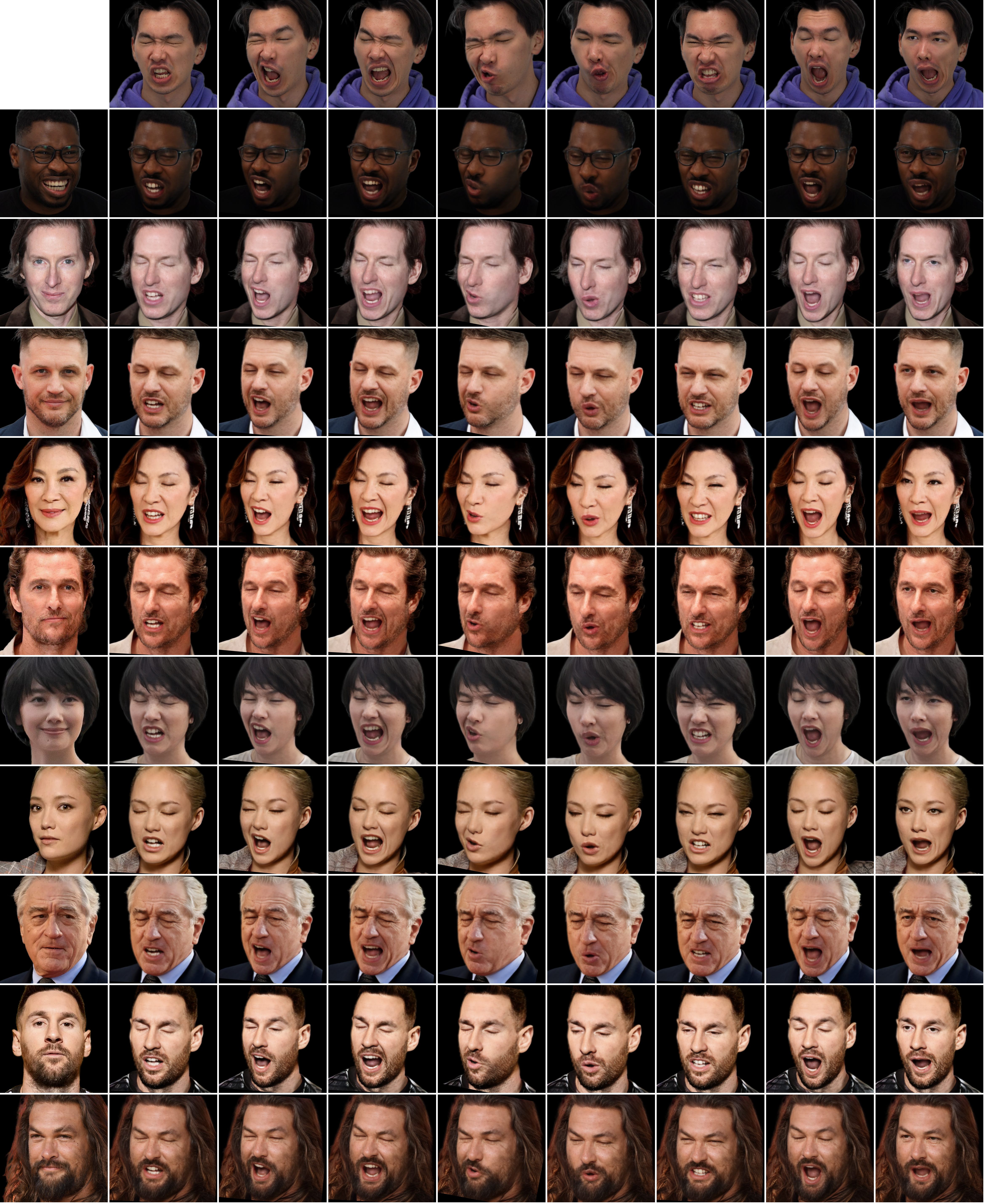}
     \caption{Additional cross-reenactment results of our method}
     \label{fig:addqual01}
 \end{figure*}

\begin{figure*}
     \centering
     \includegraphics[width=0.95\linewidth]{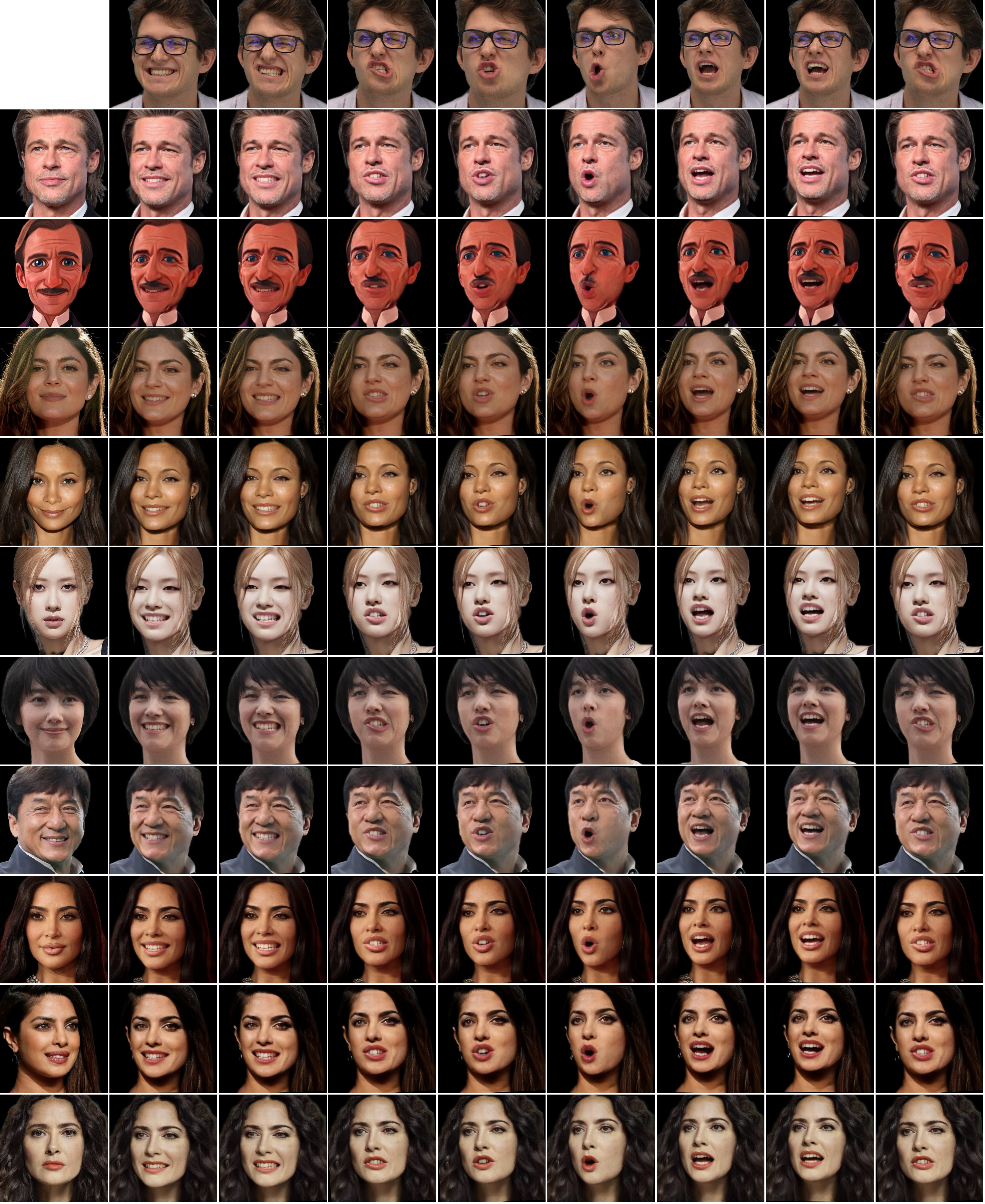}
     \caption{Additional cross-reenactment results of our method}
     \label{fig:addqual02}
 \end{figure*}

\begin{figure*}
     \centering
     \includegraphics[width=0.95\linewidth]{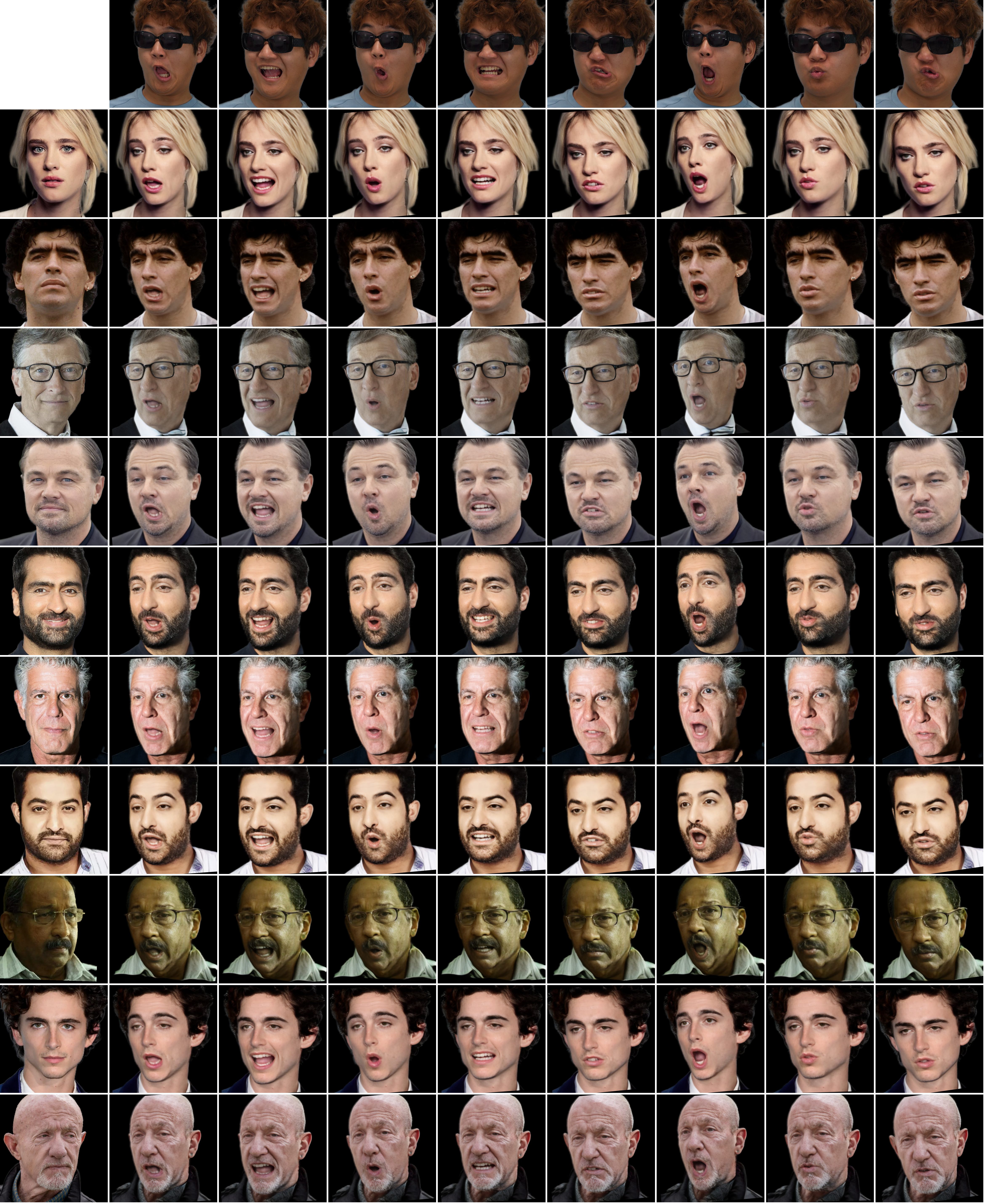}
     \caption{Additional cross-reenactment results of our method}
     \label{fig:addqual03}
 \end{figure*}

We provide 240 additional cross-reenactment results of our method in \cref{fig:addqual01}, \cref{fig:addqual02}, and \cref{fig:addqual03}. For each figure, the first column consists of the sources and the first row are the drivers. Each cell $(i, j)$ is the cross-reenactment result of the $i^{th}$ source and $j^{th}$ driver.

\bibliographystyle{ACM-Reference-Format}
\bibliography{samples/sample-base}